%% file: main.tex
\documentclass[acmlarge]{acmart}
\usepackage{graphicx}
\usepackage{amsmath}
\usepackage{float}
\usepackage{natbib}
\usepackage{tikz}
\usepackage{array}
\usepackage{xcolor, colortbl,soul}
\usepackage{graphicx}
\usepackage{tikz}
\usepackage{forest}
\usetikzlibrary{trees,positioning,shapes,shadows,arrows.meta}
\usetikzlibrary{shapes, arrows}
\usepackage{fdsymbol}

\usepackage{pgfplots}
\pgfplotsset{width=10cm,compat=1.17}
\graphicspath{ {./img/} }
\title{A Comprehensive Survey of Document-level Relation Extraction (2016-2023)}

\author{Julien DELAUNAY}
\orcid{0009-0001-9247-5745}
\email{julien.delaunay@univ-lr.fr}
\affiliation{
	\institution{Univ. La Rochelle, L3i}
	\city{La Rochelle}
	\country{France}
}
\author{Hanh Thi Hong TRAN}
\orcid{0000-0002-5993-1630}
\affiliation{
	\institution{Univ. La Rochelle, L3i}
	\city{La Rochelle}
	\country{France}
}
\author{Carlos-Emiliano GONZÁLEZ-GALLARDO}
\orcid{0000-0002-0787-2990}
\affiliation{
	\institution{Univ. La Rochelle, L3i}
	\city{La Rochelle}
	\country{France}
}
\author{Georgeta BORDEA}
\orcid{0000-0001-9921-8234}
\affiliation{
	\institution{Univ. La Rochelle, L3i}
	\city{La Rochelle}
	\country{France}
}
\author{Nicolas SIDERE}
\affiliation{
	\institution{Univ. La Rochelle, L3i}
	\city{La Rochelle}
	\country{France}
}
\author{Antoine DOUCET}
\orcid{0000-0001-6160-3356}
\affiliation{
	\institution{Univ. La Rochelle, L3i}
	\city{La Rochelle}
	\country{France}
}

\input{css_code.tex}


\begin{document}
	
	\input{sections/abstract.tex}
	
	\maketitle

	\input{sections/introduction.tex}

 \input{sections/previous_surveys}
	
	\input{sections/problem_statement.tex}

	\input{sections/methodology.tex}
	
	\input{sections/datasets.tex}

	\input{sections/methods.tex}

	\input{sections/results.tex}

 \input{sections/limitations}

	\input{sections/conclusion.tex}

\bibliographystyle{ACM-Reference-Format}
\bibliography{main}

\newpage
\appendix

\input{sections/annexe.tex}

\end{document}

%% file: css_code.tex
\ccsdesc[500]{Computing methodologies}
\ccsdesc[500]{Computing methodologies~Natural language processing}

%% file: sections/abstract.tex
\begin{abstract}
	Document-level relation extraction (DocRE) is an active area of research in natural language processing (NLP) concerned with identifying and extracting relationships between entities beyond sentence boundaries. 
	Compared to the more traditional sentence-level relation extraction, DocRE provides a broader context for analysis and is more challenging because it involves identifying relationships that may span multiple sentences or paragraphs. This task has gained increased interest as a viable solution to build and populate knowledge bases automatically from unstructured large-scale documents (e.g., scientific papers, legal contracts, or news articles), in order to have a better understanding of relationships between entities. This paper aims to provide a comprehensive overview of recent advances in this field, highlighting its different applications in comparison to sentence-level relation extraction.\\
\end{abstract}

%% file: sections/introduction.tex
\section{Introduction}

Relation extraction (RE) is a main task in natural language processing (NLP) applications that aims to automatically identify semantic relationships between entities. Its importance is showcased by its numerous applications in information extraction (IE) and knowledge base (KB) construction \cite{10.1145/3159652.3162011,trisedya-etal-2019-neural,ZAPOROJETS2021102563,prieur2023evaluating}, and in particular in the medical domain \cite{bc5cdr}. Given the unit of analysis, RE can be divided into sentence-level and document-level.

\begin{figure}[h!]
	\begin{center}
		\textit{$<$e1$>$``\textbf{The Anomaly}''$<$/e1$>$ is a novel written by $<$e2$>$\textbf{Hervé Le Tellier}$<$/e2$>$.}\\
		
		\begin{tikzpicture}[node distance = 4cm]
			\tikzstyle{person} = [rectangle, draw, text centered, rounded corners, minimum height=2em]
			\tikzstyle{album} = [rectangle, draw, text centered, minimum height=2em]
			\tikzstyle{connector} = [draw, -latex']
			
			\node [person] (e2) {Hervé Le Tellier};
			\node [person, left of=e2] (e1) {The Anomaly};
			\path [connector] (e2) -- (e1) node [midway, below] (TextNode1) {\texttt{wrote}};
			
		\end{tikzpicture}
		
	\end{center}
	
	\caption{An example of sentence-level relation extraction. In this sentence, we see that there is a relation between the book ``\textit{The Anomaly}'' and the writer \textit{Hervé Le Tellier}, which reflects the fact that the latter is the author of the first.}
	\label{fig:sentence-level_example}
\end{figure}
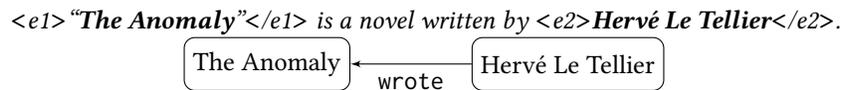

The goal of sentence-level RE is to identify the relationship between two entities that are mentioned together in the same sentence. This is a complex task because it calls for comprehension of both the sentence's content and the relationships between the entities which are often already marked with special tokens (e.g., ``$< $e$_n$$>$ [...] $<$/e$_n$$>$''). Moreover, a joint task of both named entity recognition (NER) and RE can be performed, where the same model will mark the entities and classify them, and identify the relations between each of them \cite{luan-etal-2019-general,wadden-etal-2019-entity}. 
The most common scenario is a binary relation, that can be defined as a triplet $t=<\varepsilon_i, \varepsilon_j, r>$, where $\varepsilon_i$ and $\varepsilon_j$ are the two entities, with $\varepsilon_i$ the head of the relation, $\varepsilon_j$ the tail of the relation, and $r$ the label of the relation. This kind of relation is exemplified in Figure \ref{fig:sentence-level_example}. When the relation involves more than two entities, we commonly refer to it as $n$-ary with $n \geq 2$.

As shown in Figure \ref{fig:bar}, new techniques based on graph neural networks \cite{peng-etal-2017-cross,christopoulou-etal-2018-walk,christopoulou-etal-2019-connecting}, transformers techniques \cite{wang2019finetune,tang2020hin,zhou2020documentlevel} and large-scale datasets \cite{bc5cdr,yao-etal-2019-docred,ZAPOROJETS2021102563}, made possible a progressive shift from sentence-level to document-level relation extraction (DocRE) beginning with the second-half of 2010s.

\input{tables/bar_chart}

The focus is now on identifying and classifying relations between entities across entire documents, or at least beyond the sentence boundary.  This difference in scope implies major distinctions from one task to another \cite{han2020}:
\begin{itemize}
	\item \textbf{Larger number of potential relations} The number of entities contained in a document is far greater than in a sentence, which often contains only two entities. At the scale of a full document, we have to identify relationships between all the different entities.
	\item \textbf{Handling coreferences} An entity usually occurs only once in a sentence, while in a document it can appear several times and in numerous guises. For example, in Figure \ref{fig:document-level_example}, ``\textit{He}'' is a co-referent of ``\textit{Marcus Miller}''.
	\item \textbf{Transient relationships} At the scope of a whole document, some relations require logical inference across sentences. For example, \citet{yao-etal-2019-docred} reported that approximately 40.7\% of the relations contained in DocRED, the largest dataset for DocRE, can only be extracted from reasoning over multiple sentences, even if \citet{huang-etal-2021-three} argued that in most cases, only three sentences at most are required as supporting evidence to identify a relation. In addition, this work states that importance differs widely between sentences. Sentences that do not contain any valuable information or relation actually hinder document comprehension.
\end{itemize}

\begin{figure}[h!]
	
	\textit{$<$e1$>$\textbf{Marcus Miller}$<$/e1$>$ is an American musician, songwriter, and record producer. He is best known for his work as a bassist. Main producer on the famous $<$e2$>$\textbf{Miles Davis}$<$/e2$>$' album $<$e3$>$\textbf{Tutu}$<$/e3$>$, he has also worked with pianist $<$e6$>$\textbf{Herbie Hancock}$<$/e6$>$ and saxophonist $<$e7$>$\textbf{David Sanborn}$<$/e7$>$, among others.}\\
	
	\begin{center}
		\begin{tikzpicture}[node distance = 2cm]
			\tikzstyle{person} = [rectangle, draw, text centered, rounded corners, minimum height=2em]
			\tikzstyle{album} = [rectangle, draw, text centered, minimum height=2em]
			\tikzstyle{connector} = [draw, -latex']
			
			\node [person] (e1) {Marcus Miller};
			\node [person, above of=e1] (e2) {Miles Davis};
			\node [person, left of=e1, node distance = 5cm] (e4) {Herbie Hancock};
			\node [person,  below of=e1] (e5) {David Sanborn};
			\node [album, right of=e2, node distance = 4cm] (e3) {Tutu};
			
			\path [connector] (e1) -- (e2) node [midway] (TextNode1) {\texttt{worked\_with}};
			\path [connector] (e2) -- (e1);
			\path [connector] (e1) -- (e4) node [midway, below] (TextNode2) {\texttt{worked\_with}};
			\path [connector] (e4) -- (e1);
			\path [connector] (e1) -- (e5) node [midway] (TextNode3) {\texttt{worked\_with}};
			\path [connector] (e5) -- (e1);
			\path [connector] (e1) -- (e3) node [midway, below, sloped] (TextNode4) {\texttt{produced}};
			\path [connector] (e2) -- (e3) node [midway, below] (TextNode5) {\texttt{released}};

		\end{tikzpicture}
	\end{center}
	
	\caption{An example of document-level relation extraction.}
	\label{fig:document-level_example}
\end{figure}

For all these reasons DocRE is more challenging than sentence-level RE, but there are many use cases using Knowledge Bases that require this shift such as financial investments \cite{abu2021domain} and maritime security \cite{prieur2023k}, but also teaching and classroom resources, cybersecurity, telecommunication, chemistry, geology. This survey aims to cover the following aspects:
\begin{enumerate}
	\item A comprehensive review of resources for DocRE from the past decade, including silver-standard as well as gold-standard annotated corpora.
	\item A systematic review of DocRE approaches, with a special focus on graph- and transformer-based models.
	\item A detailed discussion of the main evaluation metrics used in RE.
\end{enumerate}

The survey is organized in a subsequent manner. Section \ref{section:previous_surveys} lists relevant previous surveys related to our problem. In Section \ref{section:problem_statement}, we present the problem statement of DocRE. Section \ref{section:methodology} delineates our methodology for selecting articles to be included in the survey. The silver and gold annotated datasets for DocRE are discussed in Section \ref{section:docre_dataset}. Section \ref{section:supervised_approaches} presents a comprehensive overview of existing approaches to DocRE, while in Section \ref{section:results}, a comparative evaluation of various systems is conducted. Finally, we discuss limitations and perspectives for future work in Section~\ref{section:limitations} and conclude in Section \ref{section:conclusion}.

%% file: tables/bar_chart.tex
\begin{figure}[h]
    \centering
    \begin{tikzpicture}
        \begin{axis}
        [
        ybar,
        bar width=15pt,
        symbolic x coords={2016, 2017, 2018, 2019, 2020, 2021, 2022},
        xlabel={Year},
        ylabel={Number of Publications},
        nodes near coords,
        nodes near coords align={vertical},
        ]
        \addplot coordinates {
            (2016,13)
            (2017,6)
            (2018,19)
            (2019,28)
            (2020,54)
            (2021,82)
            (2022,99)
        };
        \end{axis}
    \end{tikzpicture}
    \caption{Number of publications per year on DocRE according to Scopus, for the terms ``document-level relation extraction'', ``cross-sentence relation extraction'', and ``inter-sentence relation extraction'' in title, abstract, or keywords. Please note that the year 2023 is not included because it had not ended at the time of submission (it will be added in case of acceptance).}
    \label{fig:bar}
\end{figure}

%% file: sections/previous_surveys.tex
\section{Previous surveys}
\label{section:previous_surveys}

Existing surveys on RE mainly focus on the more tractable task of sentence-level relation extraction. \citet{bach2007review} provide a comprehensive overview of supervised methods based on features and kernels, along with three semi-supervised approaches. Additionally, they present practical applications of RE, such as question answering and biotext mining.
\citet{de2013review} offer an exhaustive definition of the RE task, with a particular emphasis on sentence-level RE datasets and systems specialized in the Portuguese language.

Shifting the focus to the biomedical domain, \citet{cohen2005survey,zhou2014biomedical,boudjellal2020biomedical} dedicate their surveys to this area. The former presents an overview of early text mining works in the biomedical domain, encompassing sentence-level RE. The second delivers a detailed review of both binary and n-ary RE methods tailored to the biomedical domain. The latter focuses on the use of distant supervision for extracting relations from the corpus of the biological domain. Notably, it reviews a few inter-sentence and document-level methods. 

\citet{pawar2017relation} conduct an exhaustive review of sentence-level RE, covering both supervised and unsupervised approaches, alongside Open IE methods. Notably, they describe some early convolutional neural network (CNN) methods and mention some early approaches to cross-sentence RE.
\citet{shi2019brief} propose a concise review of distant supervision, with a particular emphasis on deep learning methods, incorporating some graph-based techniques. However, the discussion is limited to just two distantly supervised datasets.
\citet{bassignana-plank-2022-mean} present a review of scientific RE datasets, excluding those from the biomedical domain. While they make an earnest effort to mention document-level RE approaches and datasets and also address the ambiguity surrounding the RE task concerning relation identification and relation classification, the primary focus of the paper remains on sentence-level RE.

Our work aims to bridge a significant gap in the literature, focusing specifically on document-level relation extraction regardless of the domain of application.

%% file: sections/problem_statement.tex
\section{Problem statement}
\label{section:problem_statement}
In this survey, we follow the notation proposed in \citet{zhang-etal-2020-document}, defining an annotated document as $\mathcal{D} = \{\mathcal{S}_i\}^{n_s}_{i=1}$ with an entity set  $\mathcal{V} = \{\mathcal{E}_i\}^{n_e}_{{i=1}}$, where $\mathcal{S}_i = \{w_j\}^{n^i_w}_{j=1}$ is the $i$-th sentence with $n^i_w$ words $w$ and $\mathcal{E}_i = \{m_j\}^{n^{i}_m}_{j=1}$ is the $i$-th entity with $n^i_m$ entity mentions $m$.
In the sentence-level setting, an entity is only represented by one mention, i.e., $\forall i \in [1, n_e],\mathcal{E}_i = m_i$, which is not the case in the document-level setting, where an entity can be represented by multiple mentions.
To get a fixed representation of an entity, former methods were first to make an average of all the mention embeddings, i.e., $\mathcal{E}_i = \sum_{j=1}^{n^i_m} m_j / n^i_m$, or max pooling; but then shifted progressively to a softer representation represented by \textit{logsumexp}, facilitating the accumulation of weak signals from individual mentions: $\mathcal{E}_i = log\sum_{j=1}^{n^i_m} exp(m_j)$ \cite{Jia2019}. 
The goal of DocRE is thus to predict all intra- and inter-sentence relations $\mathcal{R}' \in \mathcal{R} = \{r_i\}^{n_r}_{i=1}$ between each entity pair.

%% file: sections/methodology.tex
\section{Methodology}
\label{section:methodology}
Google Scholar\footnote{\url{https://scholar.google.com/}}, Elsevier\footnote{\url{https://www.elsevier.com/fr-fr}}, ACL Anthology\footnote{\url{https://aclanthology.org/}}, and Google\footnote{\url{https://www.google.com/}} were primarily searched to identify papers relevant to the preparation of this study. The main search progress was tracked from April to May 2023. We then updated our survey with new methods until the date of submission. Our search terms covered two different available DocRE resources, that is corpora and systems. We used the following queries to collect research articles describing methods/systems or corpora: \textit{``inter-sentence relation extraction'', ``cross-sentence relation extraction'', ``document-level relation extraction'', ``DocRE'', ``global-level relation extraction'', ``full-abstract relation extraction''}. For datasets, we just added \textit{``corpora'', ``dataset'', ``corpus''} and \textit{``data''} to the previous queries. Since the domain is still relatively small, an exhaustive analysis is feasible therefore we decided not to rely on a threshold (i.e. a minimum number of citations regarding of the year of publication) to select articles for review.

%% file: sections/datasets.tex
\section{DocRE datasets}
\label{section:docre_dataset}
In this section, we review some important DocRE datasets, with a focus on English language resources that are publicly available along with detailed documentation regarding their structure and annotation process.\\

\input{tables/datasets_table}

\subsection{Medical domain datasets}

\subsubsection{Gold-standard annotations}

One of the first DocRE datasets published is the \textbf{BioCreative V CDR Task Corpus} (\textbf{BC5CDR (CDR)}) \cite{bc5cdr}\footnote{\url{https://biocreative.bioinformatics.udel.edu/tasks/biocreative-v/track-3-cdr/}}. Focused on Chemical Disease Relation (CDR) extraction, it consists of 1,500 PubMed articles, equally divided into sets of 500 articles for training, validation, and testing. Each entity annotation includes both the mention text spans and normalized concept identifiers, using Medical Subject Headings (MeSH) \cite{lipscomb2000medical} as the controlled vocabulary. The annotators were part of a team of four MeSH indexers for disease/chemical entity annotation and three Comparative Toxicogenomics Database (CTD) \cite{davis2009comparative} curators for Chemical-Induced Diseases (CID) relation annotation. Entities are first recorded individually by two annotators and subsequently annotated by consensus. Text-mined disease and chemical results are pre-annotated using automatic systems and displayed to the annotators to accelerate manual annotation. Relations are of two types, either ``putative mechanistic'' (the chemical influences the disease), or ``biomarker'' (the chemical correlates	with the disease). Even if this dataset is very restricted in its use case and should not be used for general purposes, it still serves as a benchmark for a lot of models, even recent ones \cite{nan-etal-2020-reasoning,zhang2022ncdre,xiao-etal-2022-sais}, given the lack of resources in DocRE datasets.\\

\citet{Luo_2022}\footnote{\url{https://ftp.ncbi.nlm.nih.gov/pub/lu/BioRED/}} introduced \textbf{BioRED}, a biomedical DocRE dataset. They annotated 600 PubMed abstracts of 11.9 sentences on average with 6 entity types (i.e., gene, disease, chemical, variant, species, and cell line) and 8 relation types. They also differentiated new findings and known facts in their annotations, avoiding replica records into the automatic knowledge construction in biomedicine. Each article was triply annotated, thus, to accelerate entity annotation, they used previous annotations combined with automated pre-annotations. However, relations were annotated by hand and from scratch, except for those that were already annotated in CDR \cite{bc5cdr}. This dataset is still recent and has not really been used yet to benchmark models. Nonetheless, since it is gold-annotated and focuses on the complete abstracts rather than only a paragraph, we think it should be highly employed.

\subsubsection{Silver-standard annotations}

\citet{peng-etal-2017-cross} built the \textbf{Ternary Drug Interaction} dataset using distant supervision to evaluate their Graph-LSTM model, here again, because of the lack of annotated datasets. The Gene Drug Knowledge Database \cite{dienstmann2015database} and the Clinical Interpretations of Variants in Cancer knowledge bases\footnote{\url{https://civicdb.org/}} are used to automatically identify mentions of drugs, genes, and mutations on the one million full-text articles from PubMed (only a small subset contains drug-gene-mutation interactions). Then, co-occurring triples with known interactions that appear in a minimal span of text are chosen as positive examples, with a limit of up to $k\leq3$ consecutive sentences. From 59 different drug-gene interactions, 3,452 positive instances are identified. However, such a process introduces a lot of noise in the corpus, resulting in a relatively low adoption to evaluate other systems.\\

Facing a lack of a reliable DocRE dataset, \citet{sahu-etal-2019-inter}\footnote{\url{http://nactem.ac.uk/CHR/}} construct a \textbf{CHemical Reactions dataset} (\textbf{CHR}) using distant supervision to evaluate their model. Consisting of 12,094 PubMed abstracts (7,298 for train, 1,182 for validation, and 3,614 for test sets), it was built using a search engine for biomedical abstracts \cite{soto2019thalia} to obtain the biomedical named entities annotations, which were then aligned with a biomedical graph database \cite{swainston2017biochem4j} to identify the relationship between each. Since CHR is very specialized in biology but also only distantly supervised with no evaluation of the accuracy of the relations identified, it was not much used as a benchmark.\\

Another example of the lack of resources in DocRE is \citet{10.1007/978-3-030-17083-7_17}\footnote{\url{https://bitbucket.org/alexwuhkucs/gda-extraction/src/master/}}, where the researchers build the \textbf{Gene-Disease-Association dataset} (\textbf{GDA}) in order to test their model, RENET. This large dataset was constructed using distant supervision on 30,192 abstracts from MEDLINE (1,000 for the test set, the rest divided between the train and validation sets following a proportion of 80:20), a platform listing a large set of gene-disease associations collected from publicly available databases. \citet{10.1007/978-3-030-17083-7_17} precise that most of the relations were curated by hand, even if the exact proportion is unknown. The GDA dataset was used to evaluate many methods \cite{christopoulou-etal-2019-connecting,minh-tran-etal-2020-dots,nan-etal-2020-reasoning,zhou2020documentlevel,sun2022enhanced,ijcai2021p0551,xie-etal-2022-eider}, often in combination with CDR \cite{bc5cdr}. \citet{su2021renet2}\footnote{\url{https://github.com/sujunhao/RENET2}} improve the dataset by incorporating full texts in addition to abstracts, creating the \textbf{RENET2} dataset. They randomly sample 500 abstracts from GDA, which were manually re-annotated by three experts with a Biology, Bioinformatics, or Computer Science background. They introduce a new association to represent that the relation between a gene and a disease is ambiguous. They use these first 500 abstracts to train their RENET2 model (discussed in Section \ref{section:supervised_approaches}), which then extracts relations from another 500 abstracts. Then, to train the model for extracting relations from the 500 full texts, they use the distantly supervised set of annotated abstracts in the training set while putting the manually annotated one in training and validation. This dataset is not used much, despite the fact that it is the only one that provides annotations for full-text articles for gene-disease associations, even if it is still silver annotated.\\

\subsection{General domain datasets}
\subsubsection{Gold-standard annotations}
Eventually, given the lack of a large, general-purpose, gold-annotated DocRE dataset, \citet{yao-etal-2019-docred}\footnote{\url{https://github.com/thunlp/DocRED}} decided to build \textbf{DocRED}.  Containing 5,053 gold-annotated documents sampled from Wikipedia, with an average length of 196.7 words, DocRED was created to handle more challenging scenarios in DocRE (and even document-level Information Extraction), such as long documents and complex entity relations. The NER part includes the classical entity labels such as \textit{PERSON}, \textit{ORGANIZATION}, and \textit{LOCATION}; while for RE, 96 relation types were used, covering a broad range of subjects (i.e., science (33.3\%), art (11.5\%), time (8.3\%), personal life (4.2\%)) including \textit{works-for}, \textit{lives-in},  \textit{place of death}, or \textit{spouse}. The dataset was created using a hybrid approach of crowdsourcing and distant supervision, where the crowdsourced annotators were asked to identify and classify entity types, and the relation labels were derived from KGs. Entities were first annotated using distant supervision and coreferenced before being manually curated by the annotators, and then linked to their corresponding Wikidata ID (entity linking).

However, relation annotation poses a difficulty. For a number $n$ of entities, there is a quadratic order of possible relations (more exactly $2\binom{n}{2} = n(n-1)$ in case of binary relations, given that the relations are directed and that there are no self-relations for entities), and most entities pairs do not actually have any relation between them. Thus, \citet{yao-etal-2019-docred} use a recommend-revise scheme. First, distant supervision from entity linking and RE models was used to generate the relations, and then only human annotators were asked to review these relations and provide supporting evidence written in the document, resulting in approximately half of the relations reserved both for those generated with entity linking and RE models.

This dataset is particularly interesting due to its large-scale and general purpose, but also due to its complexity. Only 38.9\% of the relations can be identified using simple pattern recognition.
Of the other 61.1\% of relations, 26.6\% require logical reasoning, where the relation is indirectly established by a bridge entity, 17.6\% require coreference reasoning, and 16.6\% require common-sense reasoning. Moreover, \citet{yao-etal-2019-docred} estimate that 40.7\% of the relations can only be identified when reasoning on multiple sentences.
The researchers also included a distantly-supervised version of the dataset (\textbf{DocRED distant}) with 101,873 annotated documents, far less reliable, but used in some works for model pretraining \cite{tan-etal-2022-document,tan-etal-2022-revisiting}. \\

\citet{huang-etal-2022-recommend}\footnote{\url{https://github.com/AndrewZhe/Revisit-DocRED}} point out the fact that the recommend-revise process used in the annotation of DocRED might cause bias and false negative issues, thus they manually re-annotate 96 articles. \citet{tan-etal-2022-revisiting}\footnote{\url{https://github.com/tonytan48/Re-DocRED}} equally criticize this scheme because of the Wikidata KB scarcity, which does not provide all the relations present in the text. In addition, the RE systems used to create the dataset and should leverage this problem are not up-to-date anymore, and can not compete with the models that will be trained and evaluated on DocRED.
Furthermore, the inverse relations are missing in DocRED. To overcome these limitations, they created \textbf{Re-DocRED} by reannotating 4,053 documents from DocRED (3,053 for train, 500 for validation, and 500 for test sets), generating new relations with SOTA models (KD-DocRE \cite{tan-etal-2022-document}, DocuNET \cite{ijcai2021p0551}, and ATLOP \cite{zhou2020documentlevel}) and reviewing them manually. They also added logical rules to include the inverse and co-occurring relations. As a result, each document possesses more than two times the number of relations compared to DocRED (12.5 vs. 28.1 on average for training and 12.3 vs. 34.6 for validation).
Nonetheless, they argue that since the performance on Re-DocRED is slightly better than that on DocRED, the added triples are of comparable quality to the original DocRED data. This point can be discussed since the relations coming from logical rules are far easier for the model to find, it is thus logical that the performance increases with it. Also, we hold that it is acceptable to compare models given their performances over a fixed dataset, but not the inverse. Nevertheless, Re-DocRED is a precious resource in the domain and should be used in combination with, if not instead of, the original DocRED for evaluating the performances of new DocRE systems.\\

\citet{ZAPOROJETS2021102563}\footnote{\url{https://github.com/klimzaporojets/DWIE}} released the \textbf{Deutsche Welle corpus for Information Extraction} (\textbf{DWIE}) consisting of 802 general news articles in English, randomly selected from a corpus collected from the German broadcaster Deutsche Welle between 2002 and 2018. It was annotated at the document level for NER, coreference resolution, RE, and entity linking. In contrast to the previously dominant mention-driven approaches, the dataset takes an entity-centric approach to describe interactions and properties of conceptual entities on the level of the complete document. The annotation process is defined as a bottom-up approach, involving an exploratory pass, a schema-driven pass, and an inter-rater refinement. 

During the exploratory pass, two students and one expert annotator identify the entity mentions and freely assign them a label (labels can be overlapping, i.e., an entity can have many labels), as well as freely annotate the relations to reflect as closely as possible the meaning of the article.
Then, the schema-driven pass consists in sorting all the annotations into consistent labels, both for NER (e.g., LOCATION [gpe, regio, waterbody], PERSON [politician, sport\_player, manager], EVENT [holiday, history, protest]) and RE (e.g., ``based\_in", ``citizen\_of", ``appears\_in'' or ``directed\_by'').
These new tags are organized using multi-label types illustrating different granularities in the sub-types. Moreover, \citet{ZAPOROJETS2021102563} define some logical rules to guarantee the consistencies of the annotations, such as transitivity for example.
For inter-rater refinement, the researchers reannotate 100 documents from scratch following the already defined annotation scheme. This subset is utilized to measure the inter-annotator agreement.
In the end, the resulting dataset contains numerous annotated entities and relations and should be being used for benchmarking DocRE, and more largely Document-level Information Extraction models in combination with DocRED. However, this is not the case yet, and only a small fraction of recent methods use it to benchmark their systems \cite{yu-etal-2022-relation,han2022document}.\\

\subsubsection{Silver-standard annotations}

To address the lack of a large, general domain, annotated dataset for inter-sentence RE, \citet{mandya2018dataset} produce a dataset using distant supervision. Taking inspiration from \citet{mintz2009distant}, Freebase (a large collaborative knowledge base) is first used to provide relations based on 102 labels that are then extracted from 101,042 Wikipedia sentence pairs. To balance the dataset, a limit of 2,000 sentence pairs containing the same relation is set, retaining in total 31,970 sentence pairs for the 17 most frequent relations. To evaluate the validity of the distant supervision approach, they choose four relations, for which they verify manually 50 sentence pairs where they appear, resulting in an average of 79\% of them being correct. From this analysis, they conclude that the dataset is sufficiently accurate for training and testing purposes. However, we argue that no conclusion can be drawn from the particular to the general, as the reviewed subset represents less than 1\% of the total number of sentence pairs. Despite its size, this dataset is not much used to benchmark DocRE models, probably due to a few reasons: (1) the dataset is silver annotated when DocRED \cite{yao-etal-2019-docred}, the reference dataset for DocRE, came out the year after and was also focused on the general domain but gold-annotated, (2) the dataset is balanced, which is not the case for practical applications of DocRE, where one main difficulty is to extract less frequent relations.\\

%% file: tables/datasets_table.tex
\begin{table*}[ht]
        \scalebox{0.9}{
	\begin{tabular}{lcccccc}
		\hline
		\textbf{Datasets}  & \textbf{Domain(s) }                         &  \textbf{Annotation} & \textbf{Source} & \textbf{Type} &\textbf{Nb of documents} \\ \hline
		BC5CDR (CDR) \cite{bc5cdr}       & Biology                            & Gold             & PubMed& Abstracts  & 1,500            \\ \hline
		\begin{tabular}[c]{@{}l@{}}Ternary Drug\\ Interaction \cite{peng-etal-2017-cross} \end{tabular}      & Biology                            & Silver             &  -    & - &    -     \\ \hline
		\cite{mandya2018dataset}& General& Silver&Wikipedia & Sentence pairs& 31,970\\ \hline
		CHR \cite{sahu-etal-2019-inter}           &  Medicine                                  & Silver             &   PubMed & Abstracts  &12,094\\ \hline
		GDA \cite{10.1007/978-3-030-17083-7_17}, 2021 & Medicine &  Silver             &     PubMed& Abstracts &30,192           \\ \hline
		RENET2 (abstracts) \cite{su2021renet2}& Medicine& Silver& PubMed& Abstracts &1,000\\ \hline
		RENET2 (full texts) \cite{su2021renet2}& Medicine& Silver& PubMed& Full texts &500 \\ \hline
		DocRED \cite{yao-etal-2019-docred}       & General                            &  Gold         &  Wikipedia & Paragraphs  &  5,053\\ \hline
		DocRED distant \cite{yao-etal-2019-docred}       & General                            &  Silver         &  Wikipedia & Paragraphs &101,873            \\ \hline
		Re-DocRED \cite{tan-etal-2022-revisiting}       & General                            &  Gold         &  Wikipedia  & Paragraphs   &  4,053           \\ \hline
		DWIE \cite{ZAPOROJETS2021102563}       & General                         &  Gold            &   News &  Articles & 802 \\ \hline
		BIORED \cite{Luo_2022}       & Biomedicine                            &  Gold             &     PubMed & Abstracts & 600           \\ \hline
	\end{tabular}
 }
	\caption{List of the main DocRE datasets and their characteristics. A dataset is considered to be gold-annotated when it was at least curated entirely manually.}
\end{table*}

\begin{table*}[ht]
        \scalebox{0.9}{
	\begin{tabular}{lcccc}
		\hline
		\textbf{Datasets}  & \textbf{Nb of entity mentions}                         &  \textbf{Nb of entities} & \begin{tabular}[c]{@{}l@{}}\textbf{Nb of}\\ \textbf{positive relations}\end{tabular} \\ \hline
		BC5CDR (CDR) \cite{bc5cdr}       & 10,227  & - & 3,116                \\ \hline
     \begin{tabular}[c]{@{}l@{}}Ternary Drug\\ Interaction \cite{peng-etal-2017-cross} \end{tabular}      & - & - & 3,452\\ \hline
		\cite{mandya2018dataset}& & 31,970& 31,970\\ \hline
		CHR \cite{sahu-etal-2019-inter} &  -  & - &  32,406\\ \hline
     RENET2 (full texts) \cite{su2021renet2}& - & - & 51,642 \\ \hline
		DocRED \cite{yao-etal-2019-docred}       & - & 132,375 &  62,427               \\ \hline
		Re-DocRED \cite{tan-etal-2022-revisiting}       & - & - &  $\approx$120,539               \\ \hline
		DWIE \cite{ZAPOROJETS2021102563}       & 43,373 & 23,130 & 21,749 \\ \hline
		BIORED \cite{Luo_2022} & 20,419 & 3,869 & 6,503              \\ \hline
	\end{tabular}
 }
	\caption{Number of entity mentions, entities and positive relations for the main DocRE datasets. We did not put into the table datasets for which the information were not given.}
\end{table*}

%% file: sections/methods.tex
\section{Supervised approaches for DocRE}
\label{section:supervised_approaches}

\input{tables/taxonomy_tree}

This chapter presents recent advancements in document-level relation extraction, focusing on the use of neural models for sequence-, graph-, and transformer-based approaches. The studies discussed in this chapter propose novel techniques and frameworks to improve the extraction of relations between entities across multiple sentences in a document. A taxonomy of the main DocRE approaches can be seen in Table \ref{fig:taxonomy_tree}.

\subsection{Sequence-based approaches}
An approach for performing document-level relation extraction is to view the document as an augmented sequence and apply a sequential model derived from sentence-level relation extraction to identify relations between specific entities.

\subsubsection{First CNN-based approaches}

\citet{10.1093/database/bax024}  implement two separated models for extracting Chemical-Induced Disease relations (CID) on the CDR dataset. First, a maximum entropy system is used for the inter-sentence level, and a CNN model for the intra-sentence level. Then, both results are merged to obtain relations at the document level. During the post-processing stage, if no CID can be extracted from the abstract, some heuristic rules such as associating all chemicals mentions in the title of the paper (or the most frequent in the abstract if necessary) with all diseases mentioned in the abstract are defined.

\citet{Li2018} use \textbf{recurrent piecewise convolutional neural networks} (\textbf{RPCNN})\footnote{\url{https://github.com/wglassly/CID_ATTCNN}} for CID extraction, combining domain knowledge, piecewise strategy, attention mechanism, and multi-instance learning. The method first leverages a piecewise CNN to capture the representations of individual instances belonging to a candidate. Then, a RNN is utilized to aggregate these representations across the instances within a record, enabling the generation of a comprehensive document-level representation for the candidate.

\subsubsection{Combining RNN and CNN}
\citet{mandya2018combining} propose a combined model of long short-term memory (LSTM) and CNN that exploits word embeddings and positional embeddings for cross-sentence {\it n}-ary relation extraction. The LSTM takes as input the transformed vector representation combining word embeddings and position features, the output is fed into a CNN which extracts the most important features. This approach has the advantage of being independent of intricate syntactic features, such as dependency trees, coreference resolution, or discourse features.

For the extraction of gene-disease association, \citet{10.1007/978-3-030-17083-7_17} also combines a CNN and a RNN with gated recurrent unit (GRU) gates in their \textbf{RENET} method\footnote{\url{https://bitbucket.org/alexwuhkucs/gda-extraction/src/master/}}. Here, CNN first computes sentence representations from word representations using different filters with different widths to capture the different features. This aims to capture $n$-gram patterns surrounding the target gene and target disease, which can potentially indicate a genuine association. Then, the GRU transforms them into document representation. By ablation studies, they see that even if the CNN alone might capture some important features of the document, the extra GRU layer boosts its performance. This study was expanded later in \citet{su2021renet2}\footnote{\url{https://github.com/sujunhao/RENET2}} where a section filtering step is introduced prior to the RENET system, this removes paragraphs that do not contain any gene or disease pair information and ambiguous relations.

\subsubsection{Entity-centric method}
When at the sentence level, classifying mention tuples might be sufficient, this is not the case in the document-level setting,  where any entity can be mentioned multiple times, and applying the mention-level method would explode the combinations of mention tuples. To tackle this problem, \citet{Jia2019} propose to use an entity-centric multiscale formulation by making a single relation prediction for each entity tuple. They obtain entity representation by applying \textit{logsumexp} over all the mentioned representations of the entity.\\

Nevertheless, these models inevitably encounter challenges when it comes to managing context dependencies that span long distances and require multi-hop reasoning, which is crucial for DocRE \cite{zhang-etal-2020-document}.

\subsection{Graph-based approaches}
\subsubsection{Early graph-based DocRE models}
\begin{figure*}[h!]
	\includegraphics[scale=0.17]{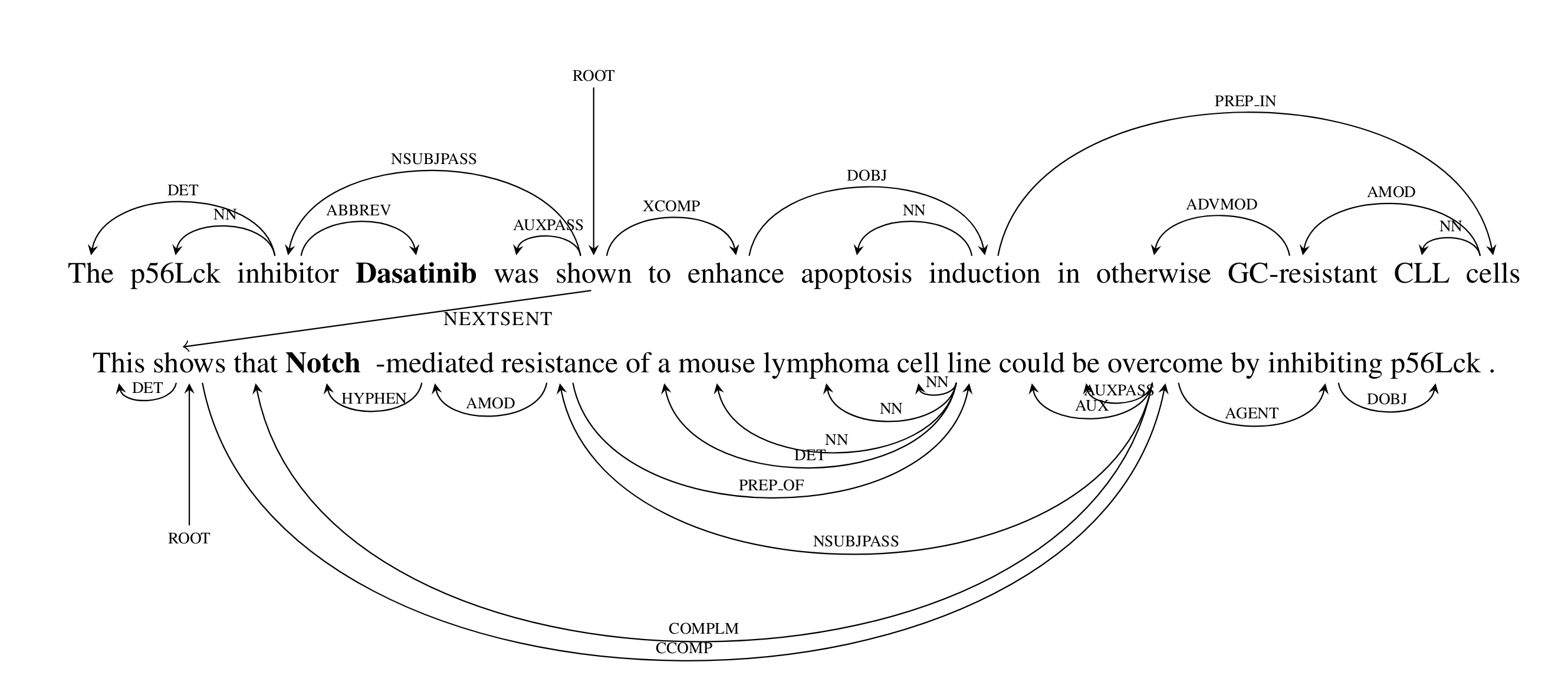}
	\caption{Document graph as described by \citet{quirk-poon-2017-distant}. Edges represent intra-sentential dependencies, as well as connections between the roots of adjacent sentences. For simplicity, the edges between adjacent words or representing discourse relations are omitted.}
	\label{fig:document_graph}
\end{figure*}

The concept of document graph was introduced by \citet{quirk-poon-2017-distant}. It provides a unifying way to derive features for classifying relations by representing words as nodes and relations (e.g., dependencies such as adjacence, syntactic dependencies, and discourse relations) as edges, see Figure \ref{fig:document_graph}. This paved the way for all the graph-based methods for DocRE. \citet{peng-etal-2017-cross} decided to apply this graph representation as a backbone to their framework, in which lies a novel graph-LSTM. In classic LSTM architectures \cite{hochreiter1997long}, each unit has only one predecessor (so only one forget gate). This is not the case for graph-LSTM, which can have multiple antecedents including connections to the same word through different edges. Thus, \citet{peng-etal-2017-cross} take inspiration from tree LSTMs \cite{tai-etal-2015-improved} and add for each unit as many forget gates as the number of precedents. Word embeddings derived from the input text are given as input of the graph-LSTM, which captures contextual representations for each word. Entities' contextual representations are combined through concatenation and used by the relation classifiers. In the case of multi-word entities, \citet{peng-etal-2017-cross} compute the average of their word representations, leaving more sophisticated aggregation methods for future research. As for backpropagation with graph-LSTM, they split the document graph into two directed acyclic graphs (DAGs). One DAG contains the left-to-right linear chain as well as other forward-pointing dependencies (forward-pass), while the other one covers the right-to-left linear chain and the backward-pointing dependencies (backward-pass). Notwithstanding, \citet{song-etal-2018-n}\footnote{\url{https://github.com/freesunshine0316/nary-grn}} argue that important information could be lost in the splitting procedure of the document into the two DAGs; thus, to keep the original graph structure as a whole, they propose a parallel state to model each word by recurrently enriching state values via message passing. By doing so, they remark that the information flow is more focused on individual words, which are exchanging information with all their graph neighbors simultaneously at each state transition. This allows the processing of the entire contextual information for extracting features for each word, compared to processing each direction of information flow separately in DAG LSTM and handling arbitrary structures such as cyclic graphs. This also permits better parallelization, speeding the computations up. However, as pointed out by \citet{verga-etal-2018-simultaneously}, the model is not very efficient due to its heavy computations, even if it paved the way for all DocRE graph-based approaches.

\begin{figure*}[h!]
	\includegraphics[scale=0.18]{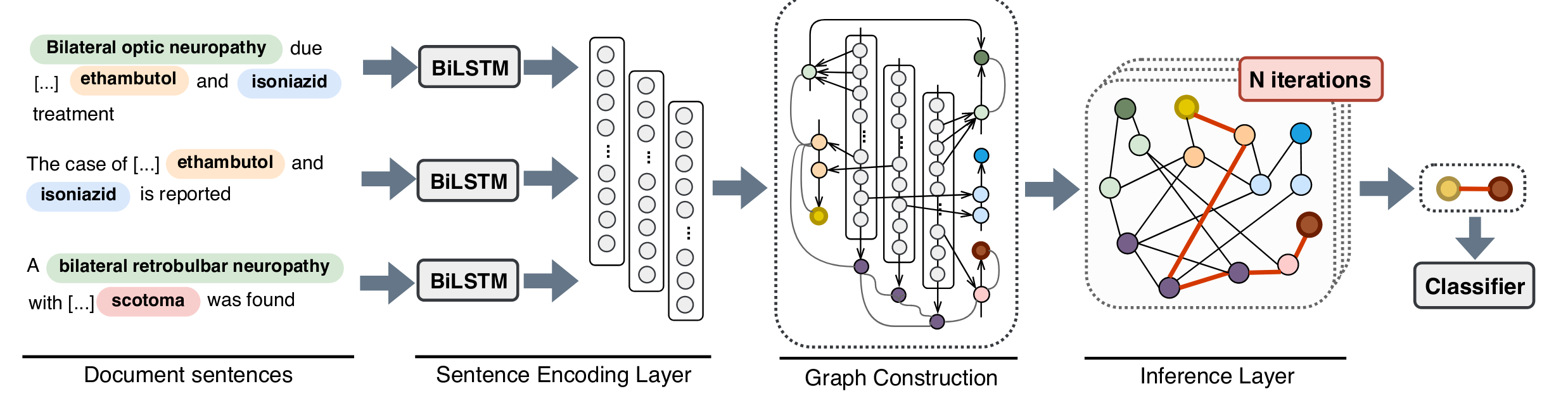}
	\caption{Architecture of the \citet{christopoulou-etal-2019-connecting} approach. The model receives a document and encodes each sentence separately. A document-level graph is constructed and fed into an iterative algorithm to generate edge representations between the target entity nodes. Some node connections are not shown for brevity (from \citet{christopoulou-etal-2019-connecting}).}
	\label{fig:christopolou}
\end{figure*}

\citet{christopoulou-etal-2018-walk} first proposed a graph walk-based neural network for sentence-level RE only, treating multiple pairs simultaneously (see Figure \ref{fig:christopolou}) and represented sentences as DAGs.
Their walk-based algorithm is composed of two steps. First, during the walk construction phase, two consecutive edges in the graph are combined using a modified bilinear transformation. Then, a walk aggregation linearly combines the initial walks and the extended walks. They later extended their approach to document-level in \citet{christopoulou-etal-2019-connecting}\footnote{\url{https://github.com/fenchri/edge-oriented-graph}}, making a big leap forward for DocRE by introducing in the \textbf{Edge-oriented Graph} (\textbf{EoG}) model a partially-connected document graph instead of the previous fully-connected one. This new document graph consists of heterogeneous types of nodes and edges in comparison with the sentence-level graph that contained only entity nodes and single edge types among them, using multi-instance learning when mention-level annotations are available. Connections between nodes are based on pre-defined document-level interactions. The objective is to generate entity-to-entity edge representations using other existing edges in the graph and consequently infer entity-to-entity relations. With this task, \citet{christopoulou-etal-2019-connecting} also found that the document-level information could help to find intra-sentence relations. Early that year, the same team proposed in \citet{sahu-etal-2019-inter} to apply a \textbf{Graph Convolutional Network (GCN)} \cite{marcheggiani-titov-2017-encoding} enabling message passing. In this approach, a document graph is constructed with syntactic dependency edge, adjacent sentence and word edges, and self-node edge but also coreferences edges to connect dependency trees of the sentence. \citet{guo-etal-2019-attention}\footnote{\url{https://github.com/Cartus/AGGCN}} also proposed to extend the GCN structure for RE by guiding them with attention (\textbf{ATGGCN}) to effectively focus on the relevant sub-graphs for the task without pruning some important information.

Relying on the dependency graph introduced in \citet{quirk-poon-2017-distant}, \citet{10.1609/aaai.v33i01.33016513}\footnote{\url{https://github.com/pgcool/Cross-sentence-Relation-Extraction-iDepNN}} employ a model that captures both the shortest and augmented dependency paths between two entities, considering both intra- and inter-sentence relations. They represent each word on a shortest-dependency path (SDP) through a dependency subtree using a combined structure of bidirectional RNNs and RecNNs  \cite{socher2012semantic}. Their proposed approach, \textbf{inter-sentential Dependency-based Neural Network} (\textbf{iDepNN}), incorporates SDP to handle inter-sentence relations by connecting the root node of a sentence's parse tree to the root of the subsequent tree. Additionally, they employ an augmented dependency path where a RecNN models the dependency subtrees for each word on the inter-sentential SDP, followed by a bidirectional RNN to extract salient semantic features. In most cases, the ADP model is better than the SDP one, however, they did not test it on the benchmark datasets for DocRE. When comparing their method with \cite{peng-etal-2017-cross}, they remark that their system is much simpler and more efficient due to its non-cyclic structures architecture.

Despite these discoveries, \citet{minh-tran-etal-2020-dots} claim that graph-based edge-oriented models only focus on the graph's edge representations, ignoring the representations of the nodes. As a result, they miss potential important clues contained in the entities and their mentions and prevent the models from leveraging the relationships between node and edge representations, as well as the similarity between representation vectors of an entity mention nodes belonging to the same entities, which could help the system performances. Therefore, they extend the EoG model from \citet{christopoulou-etal-2019-connecting} by incorporating three losses; one for Node Representation, one for Node-Edge Representation Consistency, and one for Mention Representation Consistency (\textbf{EoGANE}).

\citet{li-etal-2020-graph} noticed that since the same relation can be expressed in different sentences, and inversely that one sentence can display multiple relations, it should thus be logical to use the attention mechanism to capture the complex many-to-many interaction. They introduce \textbf{Graph Enhanced Dual Attention network} (\textbf{GEDA}), which uses a bi-directional attention mechanism consisting of the attention paid by a sentence to relation instances and the attention paid by a relation instance to sentences. By doing so, they perform better than \citet{christopoulou-etal-2019-connecting} on CDR and GDA.

\subsubsection{Information extraction in a joint setting}
The work of \citet{luan-etal-2019-general}\footnote{\url{https://github.com/luanyi/DyGIE}} focuses more largely on performing different IE tasks in a joint setting. They first proposed a sentence-level IE system, using \textbf{dynamic span graphs} (\textbf{DyGIE}) to perform NER, RE, and coreference resolution simultaneously. Since the model can be updated as new information is added to the input text, it enables the propagation of coreference and relation type confidences, which facilitates the iterative refinement of span representations. This stands in contrast to previous models, where the only interaction between tasks occurs within the shared first-layer LSTM. 
A few months later, they incorporate a cross-sentence setting into DyGIE and create \textbf{DyGIE++} \cite{wadden-etal-2019-entity}\footnote{\url{https://github.com/dwadden/dygiepp}}. The difference with DyGIE is that BERT is employed to generate token representations using a ``sliding window'' technique. Spans of text are identified and constructed by concatenating the tokens representing their left and right boundaries along with a learned span width embedding. A graph structure is dynamically generated based on the model's current estimation of the relations existing among the spans within the document. From there, equal to DyGIE, message passing facilitates the incorporation of dependencies that span across sentences. This improves the performances beyond what can be achieved by BERT alone, especially on IE tasks in specialized domains. 

\citet{WANG2021107274} use a novel rhetorical structure theory (RST)-GRAPH to build valid semantic association, as well as a set of evidence reasoning methods to select appropriate evidence to train their \textbf{discourse-aware neural RE model} (\textbf{DISCO-RE}). Training DISCO-RE jointly on RE and evidence extraction permits to alleviate the lack of ability to select and reason over evidence.

In a similar direction, \citet{xie-etal-2022-eider}\footnote{\url{https://github.com/Veronicium/Eider}} propose to empower DocRE with \textbf{Efficient Evidence Extraction and Inference-stage Fusion} (\textbf{EIDER}). During the training step, the model is jointly trained on RE and evidence extraction (labels for the latest task coming from both gold and silver annotation), while during inference, the prediction results on both the original document and the extracted evidence are combined, which encourages the model to focus on the important sentences while reducing information loss. Forcing the model to reason more along the sentences permits it to incorporate more knowledge on the relations, and thus gives EIDER higher performances.

\subsubsection{Entity-centric approaches}
\citet{ZAPOROJETS2021102563}\footnote{\url{https://github.com/klimzaporojets/DWIE}} further extend DyGIE++ with \textbf{Deutsche Welle corpus for Information Extraction} (\textbf{DWIE}) incorporating a new graph propagation technique based on attention (AttProp) which can be used in single-task as well as joint settings. Given a desire to make the model entity-centric, they cluster the mentions into their entities and consider relations between the latest. As a consequence, the graph propagation runs on the whole document. However, they only experimented on the DWIE dataset.

\begin{figure*}[h!]
	\includegraphics[scale=0.185]{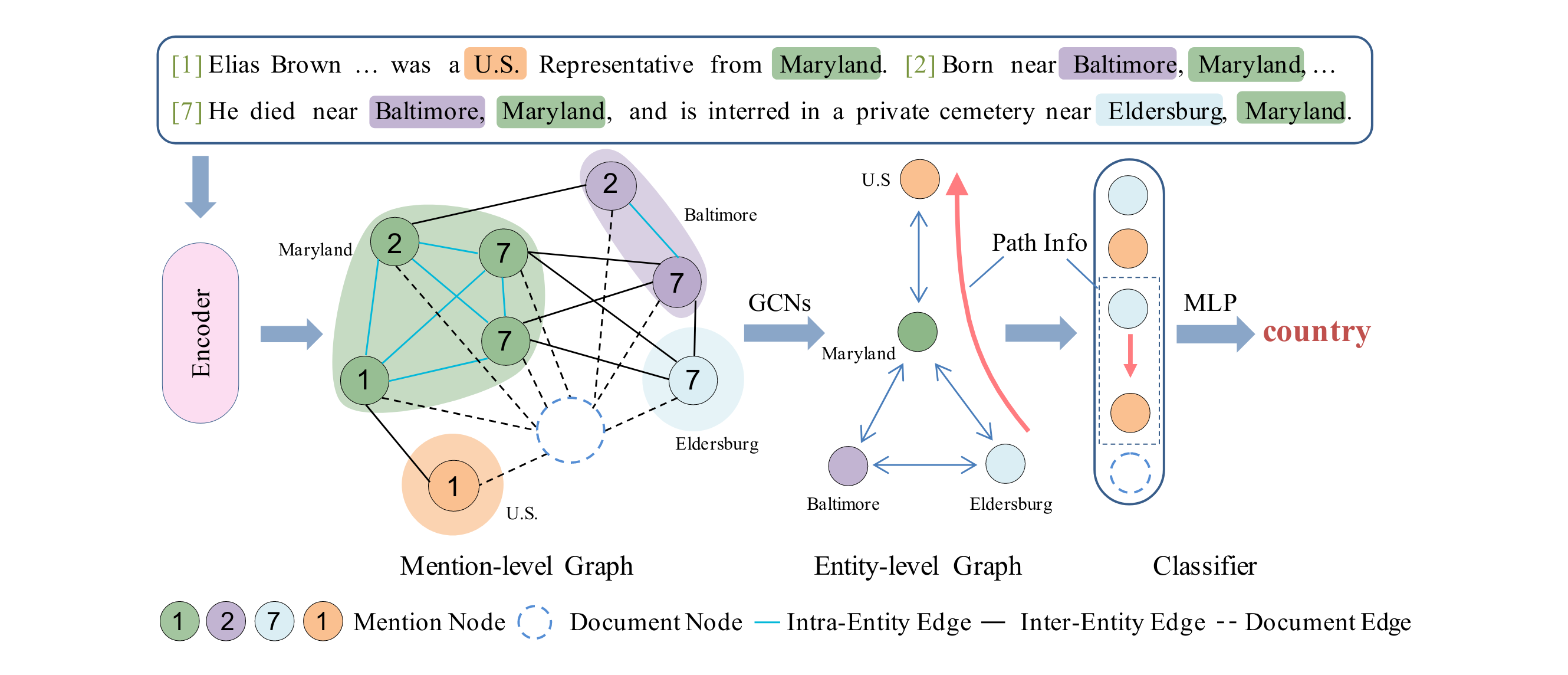}
	\caption{Architecture of GAIN \cite{zeng-etal-2020-double}. First, a context encoder obtains a contextualized representation for each word. Then, the Mention-level Graph is constructed with mention nodes and a document node. After applying GCN, an Entity-level Graph is constructed, where the paths between entities are identified for reasoning. Lastly, the classification module utilizes the gathered information to make predictions about the target relations. Different entities are in different colors. The number $i$ in the mention node denotes that it belongs to the $i$-th sentence (from \cite{zeng-etal-2020-double}).}
	\label{fig:gain}
\end{figure*}

Analogously, seeking an entity-centric solution to the DocRE problem, \citet{zeng-etal-2020-double}\footnote{\url{https://github.com/PKUnlp-icler/GAIN}} propose \textbf{Graph Aggregation-and-Inference Network} (\textbf{GAIN}) which incorporates a double graph design consisting of a heterogeneous \textit{mention-} and \textit{entity-level} graph. GAIN
captures complex interactions among different mentions and aggregates those that pertain to the same entities. This is accomplished with three types of edges: an intra-entity one to perform coreference resolution, an inter-entity one to model interactions among entities within a sentence, and a document edge, with which all mention nodes are connected to the document node.
By establishing these relations, the document node can effectively attend to all mentions and facilitate interactions between the document and individual mentions. It also addresses the challenge of long-distance dependency, since the document node can serve as a pivot for message passing using a GCN (see Figure \ref{fig:gain}). This method, in particular, showed a great improvement above SOTA, but might not be very efficient, being like other graph-based models computationally intensive. Anyway, it represents a capital step in DocRE.

\citet{xue2022corefdre} propose \textbf{Coref-aware Doc-level Relation Extraction} (\textbf{CorefDRE}) based on Graph Inference Network by instructing the model directly to learn the coreference by mention-pronoun clustering. They extract all the mentions-pronouns pairs using spaCy\footnote{\url{https://spacy.io/}} and use BERT to calculate the affinity of each mention-pronoun pair, to reduce the noise brought by pronouns. CorefDRE first constructs a Mention-Pronoun Affinity Graph (MPAG) by leveraging the mention-pronoun pairs and their affinity scores, incorporating pronoun nodes, mention nodes and various types of edges. It further applies a GCN to MPAG to capture meaningful representations for mentions and pronouns. A noise suppression mechanism merges pronouns into MPAG, transforming it into an isomorphic Entity Graph. The final step involves aggregating node and edge representations in the graph, obtaining a comprehensive representation for effective analysis and understanding.

Similar to \cite{xue2022corefdre}, \citet{LI2022109146} introduce a coreference graph that clusters expressions based on self-supervised affinity computation. These clusters serve as a node array, enabling the exposure of cross-sentence reasoning clues. They are then transformed into an \textbf{Heterogeneous Affinity Graph} (\textbf{HAG}), which combines the coreference graph with the mention graph proposed by GAIN \cite{zeng-etal-2020-double}, making it an entity graph. The HAG model shuffles the similarity to sentences containing head or tail entities using the affinity weight. To aggregate the information in the graph, HAG employs the Relational Graph Convolutional Network (R-GCN) \cite{10.1007/978-3-319-93417-4_38} and a noise suppression mechanism, resulting in a homogeneous entity graph. Even if this method permits enhancing the modeling of long-distance relations, the merging of the two graphs still introduces undesirable noise.

\subsubsection{Using a Heterogeneous graph}
To capture both sequential and structural information within documents, \citet{zhang-etal-2020-document} propose a \textbf{Dual-tier Heterogeneous Graph} (\textbf{DHG}), effectively combining these two types of information to facilitate multi-hop reasoning and final prediction. The model is composed notably of a structure modeling layer, taking into account sequence, syntax, and hierarchy for encoding the inherent structure of the document comprehensively. A relation reasoning layer facilitates multi-hop relational reasoning by propagating relational information among different entities.

However, most of these models construct relation information between all entity pairs, without knowing if there is or not a relationship between them. Thus, \citet{xu2020documentlevel}\footnote{\url{https://github.com/xwjim/DocRE-Rec}} propose to use {reconstruction} to force their \textbf{heterogeneous-based graph self-attention network} (\textbf{HeterGSAN}) to pay more attention to the learning of entity pairs with the ground-truth relations. The reconstructor generates a sequence of node representations along the path connecting two entity nodes, maximizing the probability of the path when a ground-truth relationship exists between the entity pair. This permits a further improvement in the performances of the models.

\citet{wang-etal-2020-global}\footnote{\url{https://github.com/nju-websoft/GLRE}} argue that the graph model for which \citet{christopoulou-etal-2019-connecting} opted involves irrelevant information since they are integrated indiscriminately, thus harming the system performance. Consequently, they use a \textbf{Global-to-Local Neural Networks for Relation Extraction (GLRE)} for synthesizing entity global, local, and context relation representations. BERT embeddings are utilized to capture semantic features and common-sense knowledge, and a heterogeneous graph with heuristic rules is built to model the complex interactions between all mentions, entities, and sentences in the document. The entity representations are learned more effectively by employing an R-GCN for learning global representations and aggregation of the multiple mentions of specific entities with multi-head attention for local representations. The context relation representation is finally learned using self-attention.

\citet{sun2022enhanced} propose an \textbf{Enhanced Graph Convolutional Network} (\textbf{EGCN}) model that uses information of node importance provided by PageRank \cite{page1997pagerank}.  All tokens in the document are considered nodes, which reduces the loss of semantic information. By incorporating different edge types, such as Dependency edges, Mention edges, and Entity edges,  the model is able to leverage both syntactic dependencies and semantic connections simultaneously. The importance of each node is then estimated using PageRank, which enhances the model's ability to capture topological structural information.

Later, \citet{sun2022dual} also propose to build multi-level graphs, namely token-level, mention-level, and entity-level graphs in their \textbf{Dual-Channel and Hierarchical Graph Convolutional Networks} (\textbf{DHGCN}). Moreover, they utilize a GCN to aggregate interactive information from multi-level graphs to enhance the reasoning of relations within the document. They also introduce a dual-channel module that encodes global contextual information and incorporates it into different levels of graph representation layers.

Drawing inspiration from previous works such as \citet{zeng-etal-2020-double} and \citet{zhang-etal-2020-document}, \citet{wang2023document} propose a \textbf{Multi-layer Heterogeneous Graph Neural Network} (\textbf{MHGNN}) model. It constructs a multi-layer heterogeneous graph comprising a word-level graph, a mention-level graph, and an entity-level graph. The first two graphs use a node-type and edge-type oriented heterogeneous graph attention network (NEHGAN) for message passing, respectively. Then, the entity-level graph employs the gating-based path inference mechanism to extract the multi-hop relations.

\subsubsection{Capturing context information}

\citet{zhou-etal-2020-global}\footnote{\url{https://github.com/Huiweizhou/GCGCN}} propose a \textbf{Global Context-enhanced Graph Convolutional Networks} (\textbf{GCGCN}) composed of two hierarchical inference blocks with entities as nodes and context of entity pairs as edges between nodes to capture rich global context information. A context-aware attention guided graph convolution (CAGGC) block establishes connections between two entities if they co-occur in at least one sentence. This allows CAGGC to learn a representation for an entity node by considering the contextual representations of all its mentions in the document, as well as the neighboring nodes, thereby encoding both local and global information. On the other hand, a multi-head attention guided graph convolution (MAGGC) block applies multi-head attention to generate multiple fully connected edge-weighted graphs, improving global context representations by connecting entity pairs from different sentences, and facilitating multi-hop relational reasoning.

\subsubsection{Separating intra- and inter-sentence reasoning}

\citet{kim2020} use for their \textbf{Global level Relation Extractor (GREG)} a divide and conquer approach to avoid this problem entirely by using two separate models to extract local relations and construct a knowledge graph. GREG is composed of a LSTM-based module for sentence-level RE and a knowledge graph constructor module responsible for generating knowledge graphs from the local relations; both models are trained alternatively.

\citet{li-etal-2021-mrn}\footnote{\url{https://github.com/ljynlp/MRN}} employ a locally and globally \textbf{Mention-Based Reasoning Network (MRN)} to distinguish the impacts of close and distant entity mentions. They add a new {co-predictor} to utilize both local and global entity representations for jointly reasoning the relations between close and distant entities.
The model includes a two-dimensional windowed convolution for capturing the local mention-to-mention interactions between the subject and object arguments of relations. A co-attention module is added for capturing the global interaction between each pair of entity mentions, which interact with each other to provide information between local and global contexts. The model obtained high performances, proving the interactions between intra- and inter-sentence relations.

With similar reasoning, \citet{zeng-etal-2021-sire}\footnote{\url{https://github.com/DreamInvoker/SIRE}} highlight the common practice of treating intra- and inter-sentential relations similarly in DocRE which leads to a lack of differentiation in the prediction of these distinct relation types. Not all entity pairs can be connected by a path, and some logical reasoning paths may not be fully captured in the graph, such results in many cases of logical reasoning not being covered. Their method, \textbf{SIRE}, \textbf{s}eparates \textbf{i}ntra- and inter-sentential \textbf{re}asoning. For an intra-sentential relation, they employ a sentence-level encoder to represent it in every co-occurred sentence. By aggregating the relational representations from all co-occurred sentences, they obtain a final representation that encourages intra-sentential entity pairs to emphasize the local patterns within their specific context. 
To capture inter-sentential relations, following their prior work \cite{zeng-etal-2020-double}, they adopt a document-level encoder and a mention-level graph to enable the model to capture both global and local context.
To selectively identify sentences that may signal cross-sentence relations, they employ an evidence selector, which encourages inter-sentential entity pairs to selectively focus on sentences that provide supporting evidence for their cross-sentence relations.

\citet{peng2022document}\footnote{\url{https://github.com/Crysta1ovo/SGR}} propose a \textbf{SubGraph Reasoning} (\textbf{SGR}) framework to integrate various paths into a much simpler subgraph structure to perform various reasonings at once instead of building a document-level graph first, and then focus on the overall graph structure or the paths between the target entity pair in the graph. Specifically, they construct a heterogeneous graph and devise a heuristic approach to generate reasoning paths, which capture all potential reasoning skills and encompass relevant annotated supporting sentences. This also ensures that all entity pairs can be connected by a path. Subsequently, a subgraph is extracted based on the generated paths, allowing the model to concentrate on critical entities, mentions, and sentences. Finally, a R-GCN is applied to the subgraph, enabling joint reasoning across different paths.

\subsubsection{Optimal graph structure}

\citet{nan-etal-2020-reasoning}\footnote{\url{https: //github.com/nanguoshun/LSR}} discovered that constructing a graph using co-references and rules does not always lead to the optimal graph structure. In consequence, they propose to treat the graph structure as a latent variable and induce it in an end-to-end fashion for enhanced information aggregation in the whole document. Their \textbf{Latent Structure Refinement} (\textbf{LSR}) model is comprised of a node constructor, which gives nodes as contextual representations of each mention, and a dynamic reasoner, which produces a document-level structure from the iteratively refined nodes.

\citet{huang-etal-2021-three}\footnote{\url{https://github.com/AndrewZhe/Three-Sentences-Are-All-You-Need}} focused on looking at how many sentences were required to identify a relationship in popular datasets. They found that in DocRED, when paragraphs were composed of an average of eight sentences, 96.1\% instances required no more than three sentences as supporting evidence, and 87.7\% even required only two or fewer. Similar constatations were made for CDR and GDA. One possible explanation given by the researchers is that, given the limitations of human working memory, the nature of the RE task is naturally constrained to a limited number of entities and contextual information. As described by \citet{barreyro2012working}, working memory is particularly important during reasoning tasks, but its capacity to separate information chunks is often limited to around four sentences \cite{cowan2001magical}. Therefore, the common pattern of three-sentence information chunks observed in the data may be a subconscious strategy we employ for mutual understanding. Using these assumptions, \citet{huang-etal-2021-three} designed three heuristic rules to extract paths from each document that aim to approximate supporting evidence: consecutive paths (head and tail entities are within three consecutive sentences), multi-hop paths (head and tail in distant sentences, but can be bridged via a pivot entity which co-occurs with the head and the tail in different sentences), and default paths (all the pairs of sentences where one contains the head and the other the tail). Even if their method seems simplistic, they obtained better results by combining it with a BLSTM than some SOTA methods when the paper was published \cite{nan-etal-2020-reasoning,zeng-etal-2020-double}, showing that an explicit selection of the optimal paths is better than fully relying on the graph-based model.

Previous studies have utilized heuristic rules \cite{christopoulou-etal-2019-connecting,zeng-etal-2020-double} or syntactic dependency paths \cite{sahu-etal-2019-inter, guo-etal-2019-attention} to construct a pseudo graph structure for their GNN models, establishing binary associations between pairs of entities based on task-independent auxiliary information. These approaches are considered pseudo-graph structures because they are based on task-independent auxiliary information and are not directly learned from the data, thus raising certain issues. Firstly, the graph construction overlooks many actual relational edges between entities. Secondly, the simplistic information accumulation based on the binary associative edges makes it harder to model relation-aware structural knowledge, resulting in noisy representations. To avoid the prior pseudo-graph structure, \citet{qi2022constgcn} propose \textbf{Constrained Transmission-based Graph Convolutional Network} (\textbf{ConstGCN}) to perform knowledge-based information propagation directly between entities without prior graph construction. Entity representations are updated in ConstGCN at each graph convolution step by aggregating knowledge-based information transmitted from neighboring entities across all relation spaces, enabling direct capture of the relation-aware structural semantics. To precisely follow relational paths for information transmission without a graph structure, \citet{qi2022constgcn} transmit scores that constrain the flow of information through uncertain relational edges, allowing the model to acquire semantic representations while preserving the original relation-aware structural information. By transmitting scores, ConstGCN enhances the accuracy and robustness of information propagation in the absence of a predefined graph structure.

\subsubsection{Entity representation}
In order to address the oversight of non-entity clue words in graph-based methods for document-level relation extraction, \citet{zhang2022ncdre} propose \textbf{Non-entity Clue information for Document-level Relation Extraction} (\textbf{NC-DRE}). NC-DRE utilizes a variant of transformer decoder instead of GNN models as the decoder component. This variant incorporates a Structured-Mask Multi-head Self-Attention mechanism, which effectively models heterogeneous document graphs and enhances the model's reasoning ability. Additionally, it introduces a decoder-to-encoder attention mechanism that captures more clue information, especially non-entity clue information, to improve the model's reasoning capability.

In 2023, \citet{liu2023document}\footnote{\url{https://github.com/UESTC-LHF/GRACR}} address the representation of the entity related to its mentions. Instead of simply averaging the mentions, they propose a graph information aggregation and logical cross-sentence reasoning network (GRACR). After building mention- and entity-level graphs, they use an attention mechanism to combine document embedding, aggregation, and inference information to help identify relations.

\subsubsection{Data imbalance}
One inherent problem to DocRE is data imbalance since most entity pairs do not possess a relation, i.e. their relation belongs to the N/A label. To alleviate this problem, \citet{sun2023document} utilize a two-stage framework based on dynamic graph attention networks (\textbf{TDGAT}). In the first stage, TDGAT employs a binary classification model to extract links between entity pairs. During the second stage, it employs these extracted links as prior knowledge to infer fine-grained relations among the entities.

\subsubsection{Others}
\citet{wei2022sagdre}\footnote{\url{https://github.com/IAmHedgehog/SagDRE}} propose a \textbf{S}equence-\textbf{A}ware \textbf{G}raph-based \textbf{D}oc\textbf{RE} model (\textbf{SagDRE}) which learns sentence-level
directional edges to capture the information flow in the document and uses the token-level sequential information to encode the shortest paths from one entity to the other. This permits further consideration and captures the original sequential information from the text. After the embedding generation and graph construction, multiple GCN and attention layers are stacked to learn feature representations from local and global perspectives. $k$ shortest paths are extracted from the graph, encoded by an LSTM, and fused using an attention layer to create a path embedding. To handle the problem of long-tailed multi-label problems, where an entity pair in a document can have multiple relations with a few relations being more prevalent than others, \citet{wei2022sagdre} utilize an adaptative margin loss, aiming to encourage separations between positive and negative classes by employing a threshold class.

Focusing on biomedical cross-sentence RE, \citet{ZHAO2021107230}\footnote{\url{https://github.com/DaveGabbie/Cross-sentence}} employ a BLSTM to model the context information, enhance long-distance context encoding through multi-head attention, and use a GNN to propagate the neighborhood information into nodes.
Another approach is BERT-GT by \citet{lai2021bertgt}. It integrates a neighbor-attention mechanism into the BERT architecture in combination with a Graph Transformer (GTN). This modified self-attention mechanism allows tokens to focus on their neighboring tokens, reducing noise and improving performance, particularly for long texts spanning multiple sentences. However, experiments are limited to the CDR \cite{bc5cdr} dataset.
During this same year \citet{SHI2021150} created \textbf{DocRE-HGNN}, which after encoding a document with a temporal convolution network to obtain token embeddings containing long-distance dependency information, incorporates a GTN into a heterogeneous GNN-based framework.

\subsection{Transformer-based approaches}

\subsubsection{Early Transformers-based models}

\citet{verga-etal-2018-simultaneously}\footnote{\url{https://github.com/patverga/bran}} introduced a \textbf{Bi-affine Relation Attention Network} (\textbf{BRAN}) to compute relations across mentions. BRAN predicts both intra- and inter-sentence relations between mentions simultaneously. To accomplish this, it first encodes token embeddings using a transformer layer. It then produces mention-pair predictions using a bi-affine operation of the form $W(Wx+b)+b$. Finally, it pools the mentions-level prediction using \textit{logsumexp} to obtain entity-level predictions. \citet{verga-etal-2018-simultaneously} mention that the system can be trained jointly for NER and RE, which improves robustness to noise and lack of gold annotation at the mention level. This system has the advantage of being more efficient at encoding long sequences than its contemporary model based on graph-LSTM \cite{peng-etal-2017-cross}.

To alleviate the data imbalance issue between positive and negative relations, \citet{wang2019finetune}\footnote{\url{https://github.com/hongwang600/DocRed}} adopt a \textbf{two-step finetuning process with BERT}. The former identifies if a relation exists between two mentions, and the latter identifies these entities using only the previously identified relation facts to train the model.

To enhance coreferential reasoning in language representation models, \citet{ye-etal-2020-coreferential}\footnote{\url{https://github.com/thunlp/CorefBERT}} tested their \textbf{CorefBERT} system on various downstream tasks, notably DocRE. Instead of relying on supervised coreference resolution data for fine-tuning, they introduce a novel pre-training task called Mention Reference Prediction which leverages repeated mentions in a passage and employs mention reference masking to predict the corresponding referents.  Additionally, they introduce a copy-based training objective to encourage context-based word selection and context-sensitive representations to facilitate coreferential reasoning.

\subsubsection{Local to global information aggregation}
To aggregate inference information from entity level to sentence level and then
to document level, \citet{tang2020hin} use a \textbf{Hierarchical Inference Network (HIN)}. Inspired by the translation constraint given in \citet{NIPS2013_1cecc7a7} which modeled a relation triplet $r(m_h, m_t)$ with $m_h + r \approx m_t$, they represent the relation between $\varepsilon_a$ and $\varepsilon_b$ such as $\varepsilon_b - \varepsilon_a \approx r$ for extending the assumption to the document-level. Even if it works, we argue about the exactitude of this assumption since in DocRE two entity pairs can have multiple relations, and one entity can have the same relation with multiple other entities.
They employ then a bi-affine layer to get a bilinear representation for the target entity pair.
The two previous transformations are applied in parallel in different subspaces, acquiring entity-level inference information.
Afterward, they apply a semantic matching method to compare the entity-level inference information with each sentence vector and use a hierarchical BiLSTM.
Finally, they re-employ the attention mechanism to identify the crucial sentence-level inference information, which contributes to the overall representation of document-level inference in a more comprehensive manner.

\citet{kuang2022keyword} argue that the relation between entity pairs can usually be inferred just through a few keywords. To take benefit from this, they use a Self-Attention Memory module in combination with a BERT-LSTM-based model to capture keyword features by looking for word embeddings with high cross-attention of entity pairs. Their model, much simple, achieves good improvements on both specialized and non-specialized datasets and can be applied to other ones.

\subsubsection{Entity representation}
\citet{han2020} utilize a \textbf{document-level entity mask method with type information} (\textbf{DEMMT}) to provide more information about the entities in DocRE. It introduces entity characteristics into BERT embeddings, enriching the representations with both contextual and entity-specific information. DEMMT first employs BERT to obtain contextualized representations of the entity mentions. It then uses a one-pass relation prediction method to process all the entity pairs at one time, so the text is only encoded once.

Adopting the same entity-centrism point of view as \citet{zeng-etal-2020-double} and \citet{ZAPOROJETS2021102563}, \citet{yu-etal-2022-relation}\footnote{\url{https://github.com/FDUyjx/RSMAN}} also account for the importance between each mention to represent an entity regarding a specific relation. They use a \textbf{R}elation-\textbf{S}pecific \textbf{M}ention \textbf{A}ttention \textbf{N}etwork (\textbf{RSMAN}) to incorporate essential semantics of each relation into prototype representations, and then calculate the attention between a candidate relation's prototype and the representations of mentions associated with the given entity. By considering these attentions, the model obtains a synthetic representation of the entity by weighted aggregation of all mention representations. This enables the model to capture information from multiple mentions in different representation spaces, resulting in flexible and relation-specific entity representations for different candidate relations. RSMAN can be used as a plug-in backbone to any RE model.

As said previously, in DocRE, one document contains more than one entity pair which can occur multiple times in the document and bear multiple possible relations. To tackle these issues, \citet{zhou2020documentlevel}\footnote{\url{https://github.com/wzhouad/ATLOP}} utilize two techniques, \textbf{Adaptive Thresholding and Localized cOntext Pooling} (\textbf{ATLOP}) over a BERT baseline. The adaptive thresholding permits deciding the best threshold for each entity pair by switching the global one for multi-label classification with a learnable entities-dependent one. The localized context pooling directly transfers attention from pre-trained language models to locate a relevant context that is useful to decide the relation. This solves the problem of using the same entity embedding for all entity pairs. ATLOP enriches the entity embedding with additional context relevant to the current entity pair, enabling a more comprehensive representation, and making a giant leap for DocRE. It is one of the most, if not the most, popular model for this task.

\subsubsection{Efficient use of distantly supervised labels}
For its part, \citet{xiao-etal-2020-denoising}\footnote{\url{https://github.com/thunlp/DSDocRE}} try to reduce the inherent noise induced in distantly supervised labels via three {pre-training tasks}: \textbf{Mention Entity Matching}, which involves  intra- and inter-document matching, the first focuses on establishing coreference connections within a document, while the other captures entity associations across documents; \textbf{Relation Detection}, which focuses on detecting the positive relations wrongly labeled as negative; and \textbf{Relational Fact Alignment}, which ensures that the model generates consistent representations for the same entity pair, even when expressed in different ways. \citet{xiao-etal-2020-denoising} also use Relation Detection to conduct a pre-denoising module to filter out negative instances before pre-training.

\citet{tan-etal-2022-document}\footnote{\url{https://github.com/tonytan48/KD-DocRE}} propose a multi-step approach to leverage distantly supervised data. Firstly, they employ an axial attention module as a feature extractor, allowing for attention to elements within two-hop logical paths and capturing the interdependency among relation triplets. Secondly, to handle class imbalance, they introduce an Adaptive Focal Loss that focuses on the long-tail classes, making them contribute more to the overall loss. Lastly, they utilize knowledge distillation to bridge the gap between annotated and distantly supervised data. It consists of a teacher model trained with a small amount of human-annotated data that generates predictions on a large amount of distantly supervised data, this serves as soft labels for pre-training a student model.

\subsubsection{Structure reasoning}
In their paper, \citet{xu2021entity} point out one problem of graph-based approaches for DocRE: they often separate the stages of context reasoning and structure reasoning due to the heterogeneity between the encoding network and graph network, leading to a lack of structure guidance for contextual representations. Thus, they argue that structural dependencies should be incorporated within the encoding network and throughout the overall system. They introduce the \textbf{Structured Self-Attention Network} (\textbf{SSAN}), where the entity structure is formulated under a unified framework.  Leveraging a novel extension of the self-attention mechanism \cite{NIPS2017_3f5ee243}, SSAN effectively models mention dependencies within its building blocks and across all network layers bottom-to-up by considering two entity structures, one for co-occurrence in the same sentence, and one for coreference. In addition to the dependencies between entity mentions, \citet{xu2021entity} also use an attention module to capture the relationship between entity mentions and intra-sentential non-entity words. Parallel to the attention module, they incorporate an additional module that models structural dependencies conditioned on the contextualized query/key representations. This design allows the model to benefit from the guidance of structural dependencies, ensuring that both contextual and structural information are effectively integrated.

\subsubsection{Joint learning}
\citet{ma-etal-2023-dreeam}\footnote{\url{https://github.com/YoumiMa/dreeam}} address two issues in learning both DocRE and Evidence Extraction: high resource consumption and lack of human-annotated data. For their \textbf{Document-level Relation Extraction with Evidence-guided Attention Mechanism} (\textbf{DREEAM}), they propose to direct an ATLOP model \cite{zhou2020documentlevel} on Evidence Extraction. Instead of using a separate classifier (like in \citet{xie-etal-2022-eider}), DREEAM supervises the computation of entity-pair-specific local context embeddings and incorporates attention weights over tokens as evidence scores. This approach allows them to introduce knowledge from evidence extraction without introducing additional trainable parameters or expensive matrix computations. To alleviate the lack of data they use a similar knowledge distillation technique to \citet{tan-etal-2022-document}.

\subsection{Other approaches}
\subsubsection{Semantic segmentation}
\begin{figure*}[h]
	\includegraphics[scale=0.185]{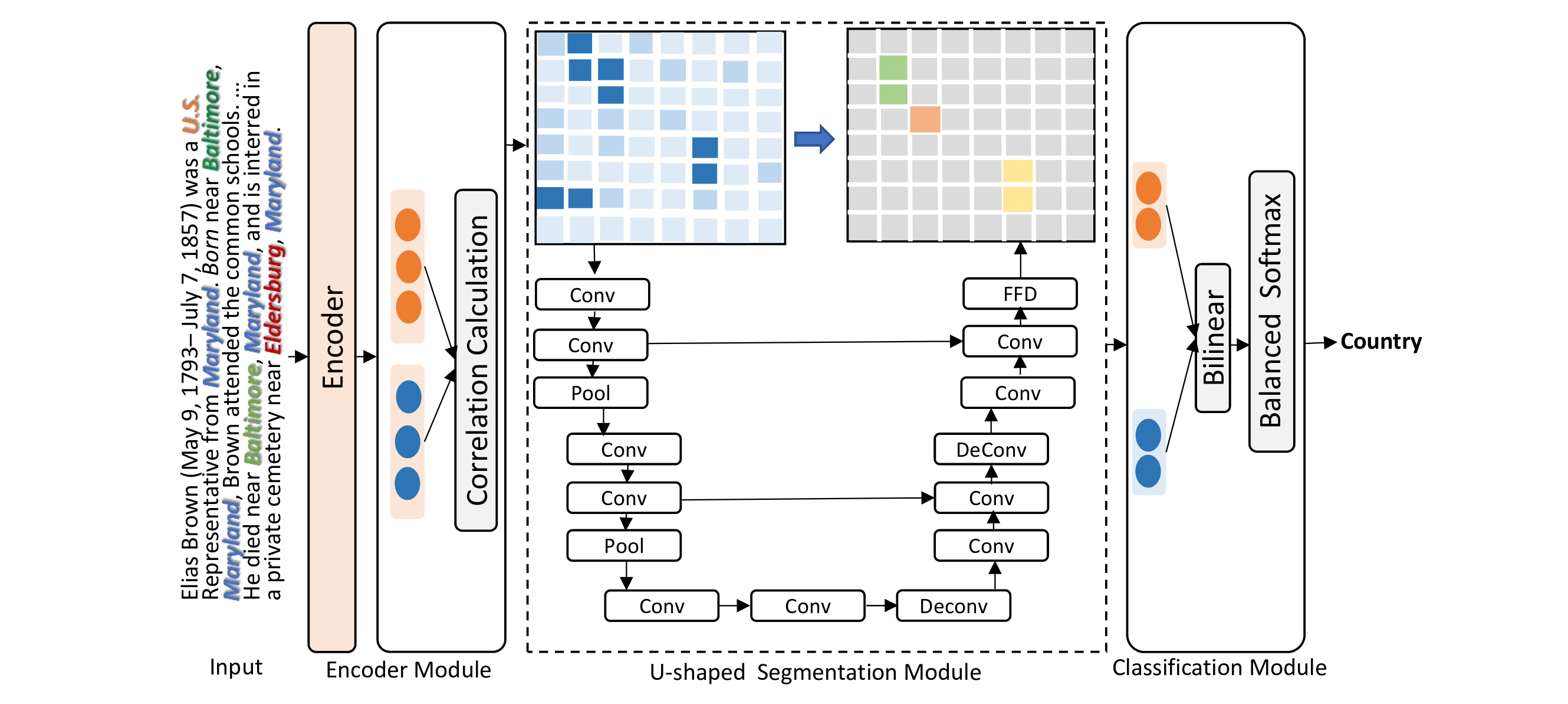}
	\caption{Architecture of DocuNet (from \cite{ijcai2021p0551}).}
	\label{fig:docunet}
\end{figure*}

\citet{ijcai2021p0551}\footnote{\url{https://github.com/zjunlp/DocuNet}} use a \textbf{Document U-shaped Network (DocuNet)} to apply a similar approach to semantic segmentation in computer vision, reformulating the task as an entity-level classification problem (see Figure \ref{fig:docunet}). Given relevant features between entity pairs as an image, the model predicts the relation type for each entity pair as a pixel-level mask. An encoder module captures the context information of entities and a U-shaped segmentation model \cite{10.1007/978-3-319-24574-4_28} captures global interdependency over the image-style feature map, a matrix containing the entity-to-entity relevance estimation. To handle the imbalance relation distribution, \citet{ijcai2021p0551} use a balanced softmax method. This original approach constitutes an important milestone in DocRE and is one of the most popular in this domain.

However, \citet{zhang2022masked} claims that capturing correlations between relations through convolutional neural networks might be unintuitive and inefficient due to the intrinsic distinction between entity-pair matrices and images. Instead, they use a \textbf{Document-level Relation Extraction model based on a Masked Image Reconstruction network} (\textbf{DRE-MIR}), which formulates the inference problem in DocRE as a masked image reconstruction problem. They adopt the DocuNet method but introduce a random masking process on the entity-pair matrix. The model then reconstructs the masked entity-pair matrix through the inference model. This approach leverages the Masked Image Reconstruction task to learn how to infer the masked points by utilizing the correlations between relations. To enhance the efficiency and effectiveness of the masked point reconstruction, they incorporate an Inference Multi-head Self-Attention module, which significantly improves the model's inference capabilities.

\subsubsection{Plug-and-play frameworks}
Because of the lack of transparency given by DocRE models, especially graph-based ones which rely heavily on implicitly powerful learned representations, \citet{ru-etal-2021-learning}\footnote{\url{https://github.com/rudongyu/LogiRE}} propose \textbf{LogiRE} to treat logic rules as latent variables. It consists of a rule generator and a relation extractor, trained simultaneously and enhancing each other using the EM algorithm.  The rule generator provides logic rules that guide the relation extractor for prediction, while the relation extractor provides supervision signals to optimize the rule generator. This approach significantly reduces the search space and allows LogiRE to be integrated as a plug-and-play technique with existing RE models.
 
 \subsubsection{Extra training tasks}
\citet{han2022document} proposed to use the interdependency of relations (i.e. relations in the real world barely appear independently, they usually interdepend on each other and appear together), which may be rich for long-tailed relations. Thus, transferring knowledge from data-rich relations to data-scarce ones might improve performance in relation extraction. To exploit this, they propose a new task, \textbf{Fined-grained Relation Co-occurrence Prediction (FRCP)}, which determines whether one relation cooccurs with a set of other relations, based on both relations and entity pairs. The learned relation embeddings which could perceive relation correlations thanks to the two auxiliary sub-tasks, accompanied by jointly-trained document contextual representations, are used to extract relational facts.
 
 To tackle the problems related to ineffective supervision and uninterpretable model predictions, \citet{xiao-etal-2022-sais}\footnote{\url{https://github.com/xiaoyuxin1002/SAIS}} propose a method called \textbf{Supervising and Augmenting Intermediate Steps} (\textbf{SAIS}). They identify coreference resolution, NER, pooled evidence retrieval, and fine-grained evidence retrieval to compose four intermediate steps involved in the reasoning process. By explicitly supervising the model's outputs in these steps and incorporating relevant contexts and entity types, \citet{xiao-etal-2022-sais} aims to improve the quality and explainability of the final outputs.
 
 \citet{xu-etal-2022-document}\footnote{\url{https://github.com/xwjim/SIEF}} focus on increasing the robustness of DocRE models: they see that most models predict correctly when an entire test document is fed as input, but display some errors when non-evidence sentences are removed. Thus, they suggest a new generic framework,  \textbf{Sentence Importance Estimation and Focusing} (\textbf{SIEF}), composed of a sentence importance score and a sentence focusing loss. The first is calculated based on the importance of each sentence in the document when it is removed from the text and is used by the second to encourage the DocRE model to focus on evidence sentences by penalizing the model for predicting relations based on non-evidence sentences.

%% file: tables/taxonomy_tree.tex
\begin{figure}
    \centering
    
\tikzset{
    basic/.style  = {draw, text width=2cm, align=center, font=\sffamily, rectangle},
    root/.style   = {basic, rounded corners=2pt, thin, align=center, fill=green!30},
    onode/.style = {basic, thin, rounded corners=2pt, align=center, fill=green!60,text width=3cm,},
    tnode/.style = {basic, thin, align=left, fill=pink!60, text width=15em, align=center},
    xnode/.style = {basic, thin, rounded corners=2pt, align=center, fill=blue!20,text width=3cm,},
    wnode/.style = {basic, thin, align=left, fill=pink!10!blue!80!red!10, text width=10em, align=center},
    edge from parent/.style={draw=black, edge from parent fork right}

}

\begin{forest} for tree={
    grow=east,
    growth parent anchor=west,
    parent anchor=east,
    child anchor=west,
    edge path={\noexpand\path[\forestoption{edge},->, >={latex}] 
         (!u.parent anchor) -- +(10pt,0pt) |-  (.child anchor) 
         \forestoption{edge label};}
}
[Supervised \\ DOCRE, basic,  l sep=6mm,
    [\small{Others}, xnode,  l sep=6mm,
        [\footnotesize{Extra training tasks}, tnode, l sep=6mm,
            [\tiny{SAIS \cite{xiao-etal-2022-sais},\\ SIEF \cite{xu-etal-2022-document}\\}, wnode]
        ]
        [\footnotesize{Plug-and-play frameworks}, tnode, l sep=6mm
            [\tiny{LogiRE \cite{ru-etal-2021-learning}}, wnode]
        ]
        [\footnotesize{Semantic segmentation}, tnode, l sep=6mm,
            [\tiny{DocuNet \cite{ijcai2021p0551},\\ DRE-MIR \cite{zhang2022masked}\\}, wnode]
        ]]
    [\small{Transformers-based approaches}, xnode,  l sep=6mm,
        [\footnotesize{Joint learning}, tnode, l sep=6mm,
            [\tiny{DREEAM \cite{ma-etal-2023-dreeam}}, wnode]
        ]
        [\footnotesize{Structure reasoning}, tnode, l sep=6mm,
            [\tiny{SSAN \cite{xu2021entity}}, wnode]
        ]
        [\footnotesize{Efficient use of \\distantly supervised labels}, tnode, l sep=6mm,
            [\tiny{\citet{tan-etal-2022-document}}, wnode]
        ]
        [\footnotesize{Multiple relations for one pair}, tnode]
        [\footnotesize{Entity representation}, tnode, l sep=6mm,
            [\tiny{DEMMT \cite{han2020},\\ ATLOP \cite{zhou2020documentlevel},\\ RSMAN \cite{yu-etal-2022-relation}\\}, wnode]
        ]
        [\footnotesize{Local to global \\information aggregation}, tnode, l sep=6mm,
            [\tiny{HIN \cite{tang2020hin}}, wnode]
        ]
        [\footnotesize{Early Transformer-based systems}, tnode, l sep=6mm,
            [\tiny{BERT-Two-Step \cite{wang2019finetune},\\ CorefBERT \cite{ye-etal-2020-coreferential}\\}, wnode]
        ] ]
    [\small{Graph-based\\ approaches}, xnode,  l sep=6mm,
        [\footnotesize{Others}, tnode, l sep=6mm,
            [\tiny{SagDRE \cite{wei2022sagdre}}, wnode]
        ]
        [\footnotesize{Optimal graph structure}, tnode, l sep=6mm,
            [\tiny{LSR \cite{nan-etal-2020-reasoning}}, wnode]
        ]
        [\footnotesize{Separating intra-\\ and inter-sentence reasoning}, tnode, l sep=6mm,
            [\tiny{MRN \cite{li-etal-2021-mrn},\\ SIRE \cite{zeng-etal-2021-sire}\\ }, wnode]
        ]
        [\footnotesize{Using a Heterogeneous graph}, tnode, l sep=6mm,
            [\tiny{DHG \cite{zhang-etal-2020-document},\\ GLRE \cite{wang-etal-2020-global},\\ HeterGSAN \cite{xu2020documentlevel},\\ EGCN \cite{sun2022enhanced}, DHGCN \cite{sun2022dual}\\}, wnode]
        ]
        [\footnotesize{Entity-centric method}, tnode, l sep=6mm,
            [\tiny{GAIN \cite{zeng-etal-2020-double},\\ CorefDRE \cite{xue2022corefdre},\\ HAG \cite{LI2022109146}\\}, wnode]
        ]
        [\footnotesize{Information extraction \\in a joint setting}, tnode, l sep=6mm,
            [\tiny{DyGIE++ \cite{wadden-etal-2019-entity},\\ EIDER \cite{xie-etal-2022-eider}\\}, wnode]
        ] 
        [\footnotesize{Early graph-based models}, tnode, l sep=6mm,
            [\tiny{GCNN \cite{sahu-etal-2019-inter},\\ EoG \cite{christopoulou-etal-2019-connecting},\\ GEDA \cite{li-etal-2020-graph}\\}, wnode]
        ]]
    [\small{Sequence-based\\ approaches}, xnode,  l sep=6mm,
        [\footnotesize{Combining RNN and CNN}, tnode, l sep=6mm,
            [\tiny{\citet{Jia2019}}, wnode]
        ],
        [\footnotesize{Entity-centric method}, tnode, l sep=6mm,
            [\tiny{\citet{mandya2018combining,10.1007/978-3-030-17083-7_17}\\}, wnode]
        ],
        [\footnotesize{First CNN-based approaches}, tnode, l sep=6mm,,
            [\tiny{\citet{10.1093/database/bax024, Li2018}\\}, wnode]
        ]
    ]
]
\end{forest}
    \caption{Taxonomy of the main DocRE methods.}
    \label{fig:taxonomy_tree}
\end{figure}

%% file: sections/results.tex
\section{Comparative results}
\label{section:results}
Although DocRE methods may vary according to the type of relationship to be extracted (gene-disease associations, generic relationships, etc.), most approaches are evaluated on DocRED, so we will use the latter to compare the results of the main methods. Comparative results for CDR and GDA datasets are given respectively on Tables \ref{table:cdr} and \ref{table:gda}. Unfortunately, since only a few methods used DWIE \cite{ZAPOROJETS2021102563} to benchmark their model yet \cite{yu-etal-2022-relation,han2022document}, we are not able to make a comparative study on this dataset. However, we insist on the fact that it should be used in the same way as DocRED for future articles, and give the available results in Appendice \ref{section:dwie}. The same can be said for BioRED \cite{Luo_2022}, and Re-DocRED \cite{tan-etal-2022-revisiting} instead of DocRED.  Since using BERT in its \textit{large} version does not affect the reasoning and comparison but only leads to better scores by using a larger model, while polluting more, all BERT-based models results are given for its \textit{base} version.\\

With respect to evaluation metrics used to provide these results,  we will consider mainly the \textit{F1 score} and \textit{IgnF1} for DocRED.
Before DocRED came out, both \textit{Precision} and \textit{Recall} were usually provided in the papers, in addition to the \textit{F1 score}. However, since 2020, there is a growing tendency to only indicate the \textit{F1 score}, and the \textit{IgnF1}, a computation of the F1 score which excludes the relations that present in both the training set and the validation and test sets.

First, we see in Table \ref{results:glove} the impact of GNN on model performances, and particularly GCN, on the domain, notably with the large improvement between the baselines \cite{yao-etal-2019-docred} and the first convolutional graphs methods \cite{sahu-etal-2019-inter,christopoulou-etal-2019-connecting}. In the same way, we can notice in Table \ref{results:bert}, the great impact of BERT-based models, even the ``simplest'' ones which can show great performances when incorporating a better understanding of the text, via coreferences \cite{ye-etal-2020-coreferential} or denoising tricks \cite{xiao-etal-2020-denoising}. We see that adopting an entity-centric point of view greatly improves performances, which can be seen in the results of GAIN \cite{zeng-etal-2020-double}.

Even if the transformer-based methods hardly compete with the graph-based ones on the front of performances, its adaptive thresholding and contextualized pooling give ATLOP \cite{zhou2020documentlevel} great results, even better than its contemporary graph-based models \cite{nan-etal-2020-reasoning}. Further improved with an efficient Evidence Extraction task \cite{ma-etal-2023-dreeam}, it can rivalize and even surpass the best graph models. This shows that, even if they perform well, the DocRE task is not constrained to graph-based solutions. The best example to illustrate this is DocuNet \cite{ijcai2021p0551}, which outperformed all of the previous models on DocRED by inspiring its solution from computer vision, only being surpassed by SIRE \cite{zeng-etal-2021-sire} in the same year. In 2022, this approach will once again give the best results with DRE-MIR \cite{zhang2022masked}.

The great performances of EIDER \cite{xie-etal-2022-eider}, NC-DRE \cite{wei2022sagdre}, and EGCN \cite{sun2022enhanced}, three graph-based models which use different ways to incorporate more knowledge in the model (EIDER by doing evidence extraction jointly with RE during training, and the other two by improving entities representations in the graph), show that making the model reasoning more over the text greatly improves the results. EGCN, SIRE \cite{zeng-etal-2021-sire}, and DHGCN \cite{sun2022dual} also show the necessity of implementing special strategies to represent both local and global information in the DocRE setting. 
 EGCN in particular shows a boom in performances (+6\% F1). Its usage of tokens as the graph's nodes, then connected with dependency, mention, and entity edges, limits the loss of information. This, combined with the great focus on nodes (thus token) importance computed by PageRank forces a great comprehension of the document, which explains this large step in performance.\\

\begin{table*}[h]
\small
	\begin{tabular}{llll}
		\hline
		\bf Model &\bf IgnF1 &\bf F1 \\ \hline
		\multicolumn{3}{c}{\bf Baselines} \\ \hline
		CNN \cite{yao-etal-2019-docred}& 36.44 & 42.33\\
		LSTM \cite{yao-etal-2019-docred}& 47.71& 50.07\\
		Bi-LSTM \cite{yao-etal-2019-docred}& 48.78& 51.06\\
		Context-Aware \cite{yao-etal-2019-docred}& 43.93& 50.64\\
		\hline
		\multicolumn{3}{c}{\bf Graph based} \\ \hline
		GCNN \cite{sahu-etal-2019-inter} * & 49.59& 51.62\\
		EoG \cite{christopoulou-etal-2019-connecting} * &49.48& 51.82\\
		GAT \cite{velickovic2018graph} * & 47.36& 49.51\\
		GEDA \cite{li-etal-2020-graph}& 51.22& 52.97\\
		GREG \cite{kim2020}& -& 52.88\\
		GCGCN-GloVe \cite{zhou-etal-2020-global}& 50.87& 53.13\\
		HIN-GloVe \cite{tang2020hin}& 51.15& 53.30\\
		LSR+GloVE  \cite{nan-etal-2020-reasoning}& 52.15& 54.18\\
		GAIN-GloVe \cite{zeng-etal-2020-double}&  52.66&       55.08\\
		HeterGSAN \cite{xu2020documentlevel}& 52.07& 53.52\\
		HeterGSAN+reconstruction \cite{xu2020documentlevel}& 53.27& 55.23\\
		Paths + BLTSM \cite{huang-etal-2021-three}& -& 56.23\\
		MRN+GloVE \cite{li-etal-2021-mrn}& 56.19& 58.46\\
		SIRE+GloVe \cite{zeng-etal-2021-sire}& 54.04& 55.96\\
		CorefDRE+GloVe \cite{xue2022corefdre}& 54.37& 56.74\\
		SGR \cite{peng2022document}& 57.15& 55.12\\
		SagDRE + GloVe \cite{wei2022sagdre}& 53.90& 56.23\\
		BLSTM+SIEF \cite{xu-etal-2022-document}& 51.03 &53.22\\
		HeterGSAN+SIEF \cite{xu-etal-2022-document}& 53.94 &55.85\\
		GAIN+SIEF \cite{xu-etal-2022-document}& 54.72 &56.75\\
		HAG+GloVe \cite{LI2022109146}& 54.37& 56.74\\
		EGCN+GloVe \cite{sun2022enhanced}& \textbf{61.26}& \textbf{63.93}\\
		DHGCN+GloVe \cite{sun2022dual}& 60.15& 61.67\\
		\hline
	\end{tabular}
	\caption{Comparative Results on the \textbf{DocRED} dataset for non BERT-based methods. All results come from their cited paper, except those with * for which it comes from \citet{nan-etal-2020-reasoning}.}
	\label{results:glove}
\end{table*}

Ultimately, DocRE is a very difficult task. Currently on DocRED, the best results are only around 65\% F1, and approximately 15\% F1
has been gained during the past four years.  Graph models, while not perfect, are more performant than transformer-based ones on this task, but less efficient. It should also be pointed out that there is only one generic benchmark dataset to assess this fact (hence the need to also use Re-DocRED and DWIE to benchmark the models). Nonetheless, the semantic segmentation approach is very promising and should be deeply investigated. Still, no model has been proposed that comprehensively addresses all the inherent challenges in DocRE.

\begin{table}[H]
\small
	\begin{tabular}{llll}
		\hline
		\bf Model &\bf IgnF1 &\bf F1 \\ \hline
		\multicolumn{3}{c}{\bf Graph based}\\ \hline
		BERT-GEDA \cite{li-etal-2020-graph}&   53.71&       55.74\\
		GCGCN-BERT \cite{zhou-etal-2020-global}& 54.53& 56.67\\
		GAIN-BERT \cite{zeng-etal-2020-double}&  \textbf{59.00}& \textbf{61.24}\\
		GAIN+SIEF \cite{xu-etal-2022-document}& 59.87 &62.29\\
		GLRE-BERT \cite{wang-etal-2020-global}& 55.40& 57.40\\
		LSR-BERT\cite{nan-etal-2020-reasoning}& 56.97& 59.05\\
		HeterGSAN-BERT \cite{xu2020documentlevel}& 56.21& 58.54\\
		HeterGSAN-BERT+reconstruction \cite{xu2020documentlevel}& 57.12& 59.45\\
		HeterGSAN-BERT+SIEF \cite{xu-etal-2022-document}& 57.93 &60.02\\
		DISCO-RE \cite{WANG2021107274}& 55.01& 55.70\\
		MRN-BERT \cite{li-etal-2021-mrn}& 59.52& 61.74\\
		SIRE-BERT \cite{zeng-etal-2021-sire}& 60.18& 62.05\\
		CorefDRE-BERT \cite{xue2022corefdre}& 60.78 &60.82\\
		EIDER-BERT \cite{xie-etal-2022-eider}& \textbf{60.42}& \textbf{62.47}\\
		NC-DRE \cite{zhang2022ncdre}& \textbf{60.59} &\textbf{62.73}\\
		SagDRE + BERT \cite{wei2022sagdre}& 60.11& 62.32\\
		HAG-BERT \cite{LI2022109146}& 60.78&  60.82\\
		EGCN-BERT \cite{sun2022enhanced}& \textbf{65.95}& \textbf{68.72}\\
		ConstGCN-BERT \cite{qi2022constgcn}& 59.58 &61.55\\
		DHGCN-BERT \cite{sun2022dual}& \textbf{65.21}& \textbf{66.87}\\
		GRACR \cite{liu2023document}& 56.47 &58.54\\
		TDGAT-BERT \cite{sun2023document}& 59.35& 61.81\\
		MHGNN \cite{wang2023document}& 59.93 &62.27\\
		\hline
		\multicolumn{3}{c}{\bf Transformer based and others}\\ \hline
		BERT \cite{wang2019finetune}& -& 53.20\\
		BERT-Two-Step \cite{wang2019finetune}& -& 53.92\\
		CorefBERT \cite{ye-etal-2020-coreferential}& 54.54&  56.96\\
		BERT+Denoising module \cite{xiao-etal-2020-denoising}& 55.53& 57.20\\
		BERT+Denoising module+Pretraining tasks \cite{xiao-etal-2020-denoising}& 56.43& 56.68\\
		BERT+DEMMT \cite{han2020}& 54.93& 57.13\\		
		ATLOP-BERT \cite{zhou2020documentlevel}&   \textbf{59.31}&       \textbf{61.30}\\
		HIN-BERT \cite{tang2020hin}& 53.70& 55.60\\
		SSAN+BERT \cite{xu2021entity}& 55.84 & 58.16\\
		DocuNet + BERT \cite{ijcai2021p0551}&   \textbf{59.93}&       \textbf{61.86}\\ 
		CorefBERT+RSMAN \cite{yu-etal-2022-relation}& 55.30& 57.53\\
		SSAN+RSMAN \cite{yu-etal-2022-relation}& 57.02& 59.29\\
		BERT+Correlation \cite{han2022document}& 59.12 (+1.01)& 61.32 (+1.12)\\
		DRE-MIR-BERT\cite{zhang2022masked}& \textbf{60.91}& \textbf{62.85}\\
		BERT-SAM \cite{kuang2022keyword}& 60.68 &62.02\\
		\cite{tan-etal-2022-document} (+BERT)& \textbf{62.56} &\textbf{64.76}\\	
		SAIS \cite{xiao-etal-2022-sais}& 60.96 &62.77\\
		BERT+SIEF \cite{xu-etal-2022-document}& 57.87 &58.93\\
       DREEAM \cite{ma-etal-2023-dreeam}& \textbf{63.73} &\textbf{65.87}\\
		\hline
	\end{tabular}
	\caption{Comparative Results on the \textbf{DocRED} dataset for BERT based methods. All results come from their cited paper, except those with * for which it comes from \citet{nan-etal-2020-reasoning}.}
	\label{results:bert}
\end{table}

\begin{table}[h]
\small
	\begin{tabular}{llll}
		\hline
		\bf Model &\bf F1 \\ \hline
        \multicolumn{2}{c}{\bf Baselines}\\ \hline
		ME-CNN \cite{10.1093/database/bax024}&   61.3\\
		BRAN \cite{verga-etal-2018-simultaneously}&   62.1\\
		RPCNN \cite{Li2018}&   59.1\\
        \hline
        \multicolumn{2}{c}{\bf Non BERT embeddings}\\ \hline
		GCNN \cite{sahu-etal-2019-inter}&   58.6\\
		EoG \cite{christopoulou-etal-2019-connecting}&   63.6\\
		EoGANE \cite{minh-tran-etal-2020-dots}&  66.1\\ 
		LSR \cite{nan-etal-2020-reasoning}& 61.2\\
		DHG-LSTM \cite{zhang-etal-2020-document}& 64.7\\ 
		MRN+BLSTM \cite{li-etal-2021-mrn}& 65.9\\
		EGCR+GloVe \cite{sun2022enhanced}& 68.2\\
		DHGCN+GloVe \cite{sun2022dual}& 73.1\\
		\hline
		\multicolumn{2}{c}{\bf BERT embeddings}\\ \hline
		DHG-BERT \cite{zhang-etal-2020-document}&   65.9 \\
		SciBERT \cite{zhou2020documentlevel}&   65.1\\
		ATLOP-SciBERT \cite{zhou2020documentlevel}&  69.4\\ 
		GLRE-BERT \cite{wang-etal-2020-global}& 68.5\\
		\cite{ZHAO2021107230}& 61.0\\
		BERT-GT \cite{lai2021bertgt}& 65.99\\
		DocRE-HGNN \cite{SHI2021150}&64.4\\
		SSAN+BERT \cite{xu2021entity}& 62.7\\
		SSAN+SciBERT \cite{xu2021entity}& 68.7\\
		SIRE+BioBERT \cite{zeng-etal-2021-sire}& 70.8\\
		DocuNet+SciBERT \cite{ijcai2021p0551}& 76.3\\
		SagDRE+SciBERT \cite{wei2022sagdre}& 71.8\\
		EIDER+SciBERT \cite{xie-etal-2022-eider}& 70.63\\
		NC-DRE+SciBERT \cite{zhang2022ncdre}& 72.05\\
		SAIS+SciBERT \cite{xiao-etal-2022-sais}& 79.0\\
		DRE-MIR-BERT\cite{zhang2022ncdre}& 76.8\\
		EGCR+BERT \cite{sun2022enhanced}& 72.5\\
		DHGCN+BERT \cite{sun2022dual}& 76.3\\
		GRACR \cite{liu2023document}& 68.8\\
		MHGNN \cite{wang2023document}& 73.04\\
		\hline
	\end{tabular}
	\caption{Comparative Results on the CDR dataset.}
    \label{table:cdr}
\end{table}

\begin{table}[h]
\small
	\begin{tabular}{llll}
		\hline
		\bf Model & \bf F1 \\ \hline
        \multicolumn{2}{c}{\bf Baselines}\\ \hline
		CNN \cite{10.1007/978-3-030-17083-7_17}& 81.3\\
		RENET \cite{10.1007/978-3-030-17083-7_17}& 83.5\\
        \hline
        \multicolumn{2}{c}{\bf Non BERT embeddings}\\ \hline
        EoG \cite{christopoulou-etal-2019-connecting}&   81.5\\
		EoGANE \cite{minh-tran-etal-2020-dots}&  82.8\\
		LSR \cite{nan-etal-2020-reasoning}& 79.6\\
		DHG-LSTM \cite{zhang-etal-2020-document}&   82.2\\
		MRN+BLSTM \cite{li-etal-2021-mrn}& 82.9\\
		EGCR+GloVe \cite{sun2022enhanced}& 83.1\\
		DHGCN+GloVe \cite{sun2022dual}& 85.6\\
        \hline
		\multicolumn{2}{c}{\bf BERT embeddings}\\ \hline
		DHG-BERT \cite{zhang-etal-2020-document}&   83.1\\
		SciBERT \cite{zhou2020documentlevel}&   82.5\\
		ATLOP-SciBERT \cite{zhou2020documentlevel}&   83.9\\
		DocRE-HGNN \cite{SHI2021150}&81.6\\
		SSAN+BERT \cite{xu2021entity}& 82.1\\
		SSAN+SciBERT \cite{xu2021entity}& 83.7\\
		SIRE+BioBERT \cite{zeng-etal-2021-sire}& 84.7\\
		DocuNet+SciBERT \cite{ijcai2021p0551}& 85.3\\
		EIDER+SciBERT \cite{xie-etal-2022-eider}& 84.54\\
		NC-DRE+SciBERT \cite{zhang2022ncdre}& 85.80\\
		SAIS+SciBERT \cite{xiao-etal-2022-sais}& 87.1\\
		DRE-MIR-BERT\cite{zhang2022ncdre}& 86.4\\
		EGCR+BERT \cite{sun2022enhanced}& 87.6\\
		DHGCN+BERT \cite{sun2022dual}& 88.2\\
		\hline
	\end{tabular}
	\caption{Comparative Results on the GDA dataset.}
    \label{table:gda}
\end{table}

%% file: sections/limitations.tex
\section{Limitations, challenges, and directions for future work}
\label{section:limitations}

Our analysis delves into the limitations of the existing body of literature surrounding the DocRE task, anticipates future challenges, and outlines potential avenues for enhancement.
DocRE is a challenging task that requires more research, as evidenced by the limited number of large-scale gold-annotated datasets and the predominant reliance on a single general-purpose dataset, DocRED \cite{yao-etal-2019-docred}, alongside two domain-specific datasets, CDR \cite{bc5cdr} and GDA \cite{10.1007/978-3-030-17083-7_17}, for benchmarking purposes. Notably, we advocate for a more extensive utilization of DWIE \cite{ZAPOROJETS2021102563} and BioRED \cite{Luo_2022}, both of which feature gold-standard annotations, as well as Re-DocRED \cite{tan-etal-2022-revisiting}, underscoring that while DocRED serves a valuable purpose, it may not still the optimal benchmark for assessing model performance.

While graph-based methods have shown promising results, they come with substantial computational overhead compared to transformer-based counterparts. Consequently, there is an enticing avenue for exploration, with large language models (LLM) emerging as a novel direction, especially in light of their cutting-edge performance across various NLP tasks \cite{brown2020language, openai2023gpt4, 10.1145/3560815}. Similarly, it would be interesting to apply SOTA methods for semantic segmentation to DocRE \cite{THISANKE2023106669}.

In addition to these technical challenges, there are also some important practical considerations for DocRE. As emphasized by \citet{boudjellal2020biomedical}, the initial step of NER bears immense significance in the DocRE pipeline. An inadequately executed NER task can severely compromise the accuracy of the entire DocRE process. Furthermore, \citet{ZAPOROJETS2021102563} have highlighted the intricate interplay between relations and entities, suggesting that joint training of NER and RE models could yield substantial benefits. Thus, a concerted effort should be directed towards advancing the joint performance of NER and RE. Moreover, the majority of existing DocRE datasets focus primarily on named entities, overlooking the potential importance of other entity types, such as terms and concepts.

%% file: sections/conclusion.tex
\section{Conclusion}
\label{section:conclusion}

As document-level relation extraction is far more difficult than its sentence-level counterpart, more research is needed to improve results. Indeed, as this survey shows the domain remains relatively shallow compared to its numerous use cases: only a few large-scale gold-annotated datasets are available and just one dataset is used to benchmark the models, despite some inherent problems. While the evolution of the solutions in the domain these last six years seems promising, we note that there is still room for improvement as the document-level setting conceals many subtleties. A possible direction for future work is to improve the efficiency of graph-based methods since they are computationally intensive but with competitive results.

Our survey compiles a comprehensive overview of the advancements made in the DocRE task from 2016 to mid 2023. We first defined the DocRE task, and provided a comprehensive list of the well-annotated datasets available in the domain, highlighting their advantages and limitations. Then, we summarized the key sequence-based, graph-based, and transformer-based approaches and made a comparison of their performance in supervised DocRE. Ultimately, it brings into focus certain constraints inherent in the proposed methodologies and puts forward recommendations for prospective endeavors.

\section{Acknowledgements}
\label{section:acknowledgements}
The first author was partly funded by Région Nouvelle-Aquitaine. This work has also been supported by the TERMITRAD (AAPR2020-2019-8510010) project funded by the Nouvelle-Aquitaine Region, France.

%% file: sections/annexe.tex
\section{Comparative results on DWIE}
\label{section:dwie}

\begin{table*}[h]
	\begin{tabular}{llll}
		\hline
		\bf Model &\bf IgnF1 &\bf F1 \\ \hline
		CNN $\spadesuit$ & 34.65 &46.14\\
		LSTM $\spadesuit$& 40.81 &52.60\\
		BLSTM $\spadesuit$& 42.03 &54.47\\
		Context-Aware $\spadesuit$& 45.37 &56.58\\
		CorefBERT $\clubsuit$& 61.71 & 66.59\\
		GAIN $\spadesuit$& 62.37 & 67.57\\
		SSAN $\clubsuit$& 62.58 &69.39\\
		ATLOP $\spadesuit$&62.09 &69.94\\
		CorefBERT+RSMAN \cite{yu-etal-2022-relation}& 62.01& 67.52\\
		SSAN+RSMAN \cite{yu-etal-2022-relation}& 63.42& 70.95\\
		BERT$_{base}$ (2022) \cite{han2022document}& 62.92& 69.12\\
		BERT+Correlation \cite{han2022document}& 65.64& 71.56\\		
		\hline
	\end{tabular}
	\caption{Comparative Results on the \textbf{DWIE} dataset \cite{ZAPOROJETS2021102563}. Results with $\spadesuit$ and $\clubsuit$ come from \citet{ru-etal-2021-learning} and \citet{yu-etal-2022-relation}, respectively.}
\end{table*}